\documentclass{article}

\usepackage{microtype}
\usepackage{afterpage}
\usepackage{graphicx}
\usepackage{lipsum}
\usepackage{subcaption}
\usepackage{booktabs} % for professional tables

\usepackage{hyperref}

\usepackage[accepted]{arxiv}

\usepackage{amsfonts}       % blackboard math symbols
\usepackage{amsmath}        % For \tfrac, etc.
\usepackage{amsthm}
\usepackage{bm}
\usepackage{color}
\usepackage{tikz}
\usetikzlibrary{arrows,shapes,backgrounds,through,shadows}

\DeclareMathOperator{\diag}{diag}
\def\KL{\mathrm{KL}}
\def\datapoint{_{(n)}}  % data point index

\def\a{\mathbf{a}}
\def\b{\mathbf{b}}
\def\c{\mathbf{c}}

\def\m{\mathbf{m}}

\def\r{\mathbf{r}}
\def\s{\mathbf{s}}
\def\t{\mathbf{t}}

\def\x{\mathbf{x}}
\def\y{\mathbf{y}}
\def\z{\mathbf{z}}

\def\0{\mathbf{0}}
\def\1{\mathbf{1}}
\def\2{\mathbf{2}}

\def\I{\mathbf{I}}

\def\R{\mathbf{R}}

\def\X{\mathbf{X}}
\def\Y{\mathbf{Y}}
\def\Z{\mathbf{Z}}

\def\balpha{\bm{\alpha}}

\def\bgamma{\bm{\gamma}}

\def\blambda{\bm{\lambda}}

\def\bmu{\bm{\mu}}

\def\bphi{\bm{\phi}}
\def\bpi{\bm{\pi}}
\def\bsigma{\bm{\sigma}}

\def\btheta{\bm{\theta}}
\def\bxi{\bm{\xi}}

\def\drm{\mathrm{d}}

\def\Ebb{\mathbb{E}}

\def\Rbb{\mathbb{R}}

\def\Bcal{\mathcal{B}}
\def\Ccal{\mathcal{C}}
\def\Dcal{\mathcal{D}}
\def\Fcal{\mathcal{F}}
\def\Gcal{\mathcal{G}}

\def\Lcal{\mathcal{L}}

\def\Ncal{\mathcal{N}}

\def\Qcal{\mathcal{Q}}

\def\Tcal{\mathcal{T}}

\def\defined{\stackrel{\text{\tiny def}}{=}}

\definecolor{Purple}{rgb}{0.63,0.1,0.76}
\definecolor{DarkGreen}{rgb}{0.1,0.7,0.2}
\definecolor{LightGray}{rgb}{0.73,0.7,0.65}
\definecolor{LightBlue}{rgb}{0.6,0.6,0.9}

\tikzstyle{disc}=[rectangle,
draw=blue!50,
thick,inner sep=0pt,
line width=1pt,
minimum size=6mm]  % discrete node
\tikzstyle{rnn}=[diamond,draw=black,thick,inner sep=0pt,line width=1pt,minimum size=8mm]  % rnn hidden node
\tikzstyle{randomv}=[circle,draw=red,thick,inner sep=0pt,minimum size=8mm,>=stealth]  % random variable  node
\tikzstyle{obs}=[fill=blue!20,thick]  % observed node
\tikzstyle{ocont}=[circle,draw=blue!50,thick,inner sep=0pt,minimum size=8mm,>=stealth]  % continuous  node
\tikzstyle{dgraph}=[->, line width=1pt]
\tikzstyle{ugraph}=[line width=1pt]

\allowdisplaybreaks

\icmltitlerunning{A Factorial Mixture Prior for Compositional Deep Generative Models}

\begin{document}

\twocolumn[
\icmltitle{A Factorial Mixture Prior for Compositional Deep Generative Models}

\vskip 0.1in

\vbox{%
\hsize\textwidth
\linewidth\hsize
\centering
\begin{tabular}[t]{c}
\textbf{Ulrich Paquet} \\ DeepMind
\end{tabular}
\hspace{25pt}
\begin{tabular}[t]{c}
\textbf{Sumedh K.~Ghaisas} \\ DeepMind
\end{tabular}
\hspace{25pt}
\begin{tabular}[t]{c}
\textbf{Olivier Tieleman} \\ DeepMind
\end{tabular}
}

\icmlkeywords{Machine Learning, ICML}

\vskip 0.4in
]


\begin{abstract}
We assume that a high-dimensional datum, like an image,
is a compositional expression of a set of properties,
with a complicated non-linear relationship between the
datum and its properties.
This paper proposes a factorial mixture prior for capturing latent properties,
thereby adding
structured compositionality to deep generative models.
The prior treats a latent vector
as belonging to Cartesian product of subspaces,
each of which is quantized separately
with a Gaussian mixture model.
Some mixture components can be set
to represent properties as observed random variables whenever labeled properties are present.
% The mixture priors share a conjugate exponential hyperprior.
Through a combination of stochastic variational inference and 
gradient descent, a method for learning how to infer discrete properties in an unsupervised or semi-supervised way
is outlined
and empirically evaluated.
\end{abstract}

\section{Introduction}
\label{sec:introduction}

An image $\x$ is often compositional in its properties or attributes.
We might generate a random portrait $\x$ by first picking discrete
properties that are simple to describe, like whether the person should be \texttt{male},
whether the person should be \texttt{smiling},
whether the person should wear \texttt{glasses},
and maybe whether the person should have one of a discrete set of \texttt{hair colors}.
In the same vein, we might include discrete properties that are harder to describe, but are still present in typical portraits. These are properties that we wish to discover in an unsupervised way from data,
and may represent background textures, skin tone, lighting, or any property that we didn't explicitly enumerate.

This paper proposes a
factorial mixture prior for capturing such properties,
and presents a method for learning how to infer discrete properties in an
unsupervised or semi-supervised way.
The outlined framework adds structured compositionality to deep generative models.

\begin{figure}[t]
\centering
% \vskip 0.2in
\begin{tabular}[t]{c@{\hskip 4pt}c@{\hskip 4pt}c@{\hskip 4pt}c@{\hskip 4pt}c}
& $k_2 = 1$ & $k_2 = 2$ & $k_2 = 1$ & $k_2 = 2$ \\
\rotatebox[origin=l]{90}{\hspace{8pt}$k_1 = 1$} &
\includegraphics[width=0.21\columnwidth]{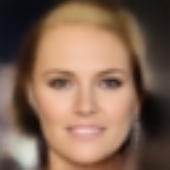} &
\includegraphics[width=0.21\columnwidth]{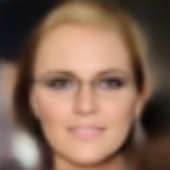} &
\includegraphics[width=0.21\columnwidth]{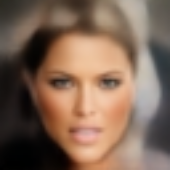} &
\includegraphics[width=0.21\columnwidth]{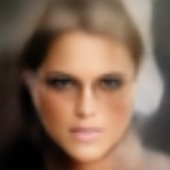}

\\
\rotatebox[origin=l]{90}{\hspace{8pt}$k_1 = 2$} &
\includegraphics[width=0.21\columnwidth]{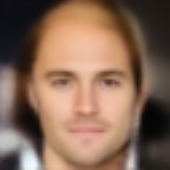} &
\includegraphics[width=0.21\columnwidth]{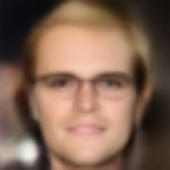} &
\includegraphics[width=0.21\columnwidth]{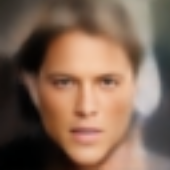} &
\includegraphics[width=0.21\columnwidth]{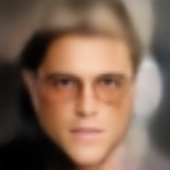}
\\
& \multicolumn{2}{c}{\hspace{-5pt}\upbracefill}
& \multicolumn{2}{c}{\hspace{-5pt}\upbracefill} \\
& \multicolumn{2}{r}{given $\z_3 \sim \z_3 | k_3 = 14\phantom{.}$}
& \multicolumn{2}{r}{given $\z_3 \sim \z_3 | k_3 = 14\phantom{.}$}
\\
& \multicolumn{2}{r}{and $\z_4 \sim \z_4 | k_4 = 27\phantom{.}$}
& \multicolumn{2}{r}{and $\z_4 \sim \z_4 | k_4 = 27\phantom{.}$}
\end{tabular}
\caption{
Samples from a deep generative model $p_{\theta}(\x | \z)$.
For the left and right blocks of faces, we sampled
$\z_3$ and $\z_4$ once,
and constructed $\z$ as a Cartesian product
$\z = [\z_1, \z_2, \z_3, \z_4]$.
The remaining $\z_1$ and $\z_2$ vectors were iteratively
filled in by sampling $\z_1 | k_1$ and $\z_2 | k_2$ for all
$(k_1, k_2)$ settings.
With background, skin tone and other properties being kept constant
in the left and right blocks,
we mark a visible change in gender in the two settings of latent variable $k_1$. The presence or absence of glasses on a portrait $\x$ is captured
by latent variable $k_2$.
Following the discussion in Section \ref{sec:semi-supervised}, 
with semi-supervision, mixture components
$(k_1 = 2, k_2 = 2)$ are responsible for rendering
$(\texttt{male}, \texttt{glasses})$ portraits.
}
\label{fig:faces-and-glasses}
% \end{center}
% \vskip -0.2in
\end{figure}

To make the example of learning to render a portrait $\x$ concrete,
let each property $i = 1, .., I$, be it one that we can describe or
not, take values $k_i \in \{ 1, .., K_i \}$.
Given a sampled latent value $k_i$,
we generate a real-valued vector
$\z_i \in \Rbb^{D}$ from a mixture component $k_i$
in a different $K_i$-component mixture model for each property $i$.
The $\z_i$-vectors are concatenated into a latent vector $\z = [\z_1, .., \z_I]$,
from which $\x$ is generated according to any deep generative model
$p_{\btheta}(\x | \z)$ that is parameterized by $\btheta$.

The parameters of the aforementioned generative process can be trained wholly unsupervised, and a factorial structure of clustered latent properties will naturally emerge if it
exists in the training data.
However, unsupervised learning provides no guarantee that some $\z_i$ encodes and models a labelled property of $\x$.
Some semi-supervised learning is required
to associate a named property with some $\z_i$.
A few training data points might be accompanied with sets of labels, an example set being $\y = \{ \texttt{female}, \texttt{no glasses} \}$.
We may choose which labels in the set to map to properties $i = 1, 2, \ldots$, and then when present in a training example,
use these labels in a semi-supervised way.

Continuing the illustrative example, let $k_1$ index the \texttt{gender}
of portrait $\x$ and $k_2$ whether \texttt{glasses} appear in $\x$ or not, to yield two
semi-supervised latent clusters with $K_1 = K_2 = 2$ for $\z_1$ and $\z_2$.
In Figure \ref{fig:faces-and-glasses},
we let factors $i = 3, 4$ be free to capture other combinations of variation in $\x$. For 
a given unsupervised cluster pair $(k_3, k_4)$ of latent properties,
Figure \ref{fig:faces-and-glasses} illustrates how
portraits $\x$ of males and females with
and without glasses
are sampled by iterating over $k_1$ and $k_2$.

Despite the naive simplicity of the generative model,
inferring the posterior distribution of $\z$ and the properties'
latent cluster assignments for a given $\x$ is far from trivial.
Next to inference stands parameter learning.
Section \ref{sec:model} is devoted to the model, inference and learning.
Section \ref{sec:related}
traces the influences on this model, and related work from
the body of literature on ``untangled'', structured latent representations.
Empirical evaluations
% in the spirit of Figure \ref{fig:faces-and-glasses}
follow in Section
\ref{sec:results}.

%!TeX root = main.tex

\section{A Structured, Factorial Mixture Model}
\label{sec:model}

We observe data $\X = \{ \x^{\datapoint} \}_{n=1}^N$,
and model each $\x^{\datapoint}$ with a likelihood
$p_{\btheta}(\x^{\datapoint} | \z^{\datapoint})$,
a deep and flexible function with parameters $\btheta$.
A few data points might have labels, and we denote their indices by set $\Omega$ and labels by
$\Y = \{ \y^{\datapoint} \}_{n \in \Omega}$.
We will ignore the labels for the time being, and return to them in
Section \ref{sec:semi-supervised} when we incorporate them as a semi-supervised signal to aid the prior factors in corresponding to named properties.
Additionally, when clear from the context, we will omit superscripts $n$.

%%%%%%%%%%%%%%%%%%%%%%%%%%%%%%%%%%%%%%%%%%%%%%%%%%%%%%%%%
\subsection{Factorial Mixture Prior on $\z^{\datapoint}$}
%%%%%%%%%%%%%%%%%%%%%%%%%%%%%%%%%%%%%%%%%%%%%%%%%%%%%%%%%

We present a prior that decomposes $\z$ in a discrete, structured way.
With a subdivision of $\z$ into sub-vectors $\z_i$,
let each factor $\z_i$ be generated by a separate Gaussian mixture model with $K_i$ components
with non-negative mixing weights $\bpi_i$ that sum to one, $\sum_{k=1}^{K_i} \pi_{ik} = 1$.
Let factor $i$'s component 
$k$ have a mean $\bmu_{ik}$ and diagonal precision matrix
$\diag(\balpha_{ik})$,
and define
$\bxi = \{ \bpi_i, \{ \bmu_{ik}, \balpha_{ik} \}_{k=1}^{K_i} \}_{i=1}^I$ to denote all prior parameters as random variables, so that
$
p(\z | \bxi) = \prod_i \sum_k \pi_{ik}\, \Ncal(\z_i ; \bmu_{ik}, \diag(\balpha_{ik})^{-1})
$.
Each $\z_i$ is augmented
with a cluster assignment vector $\r_i \in \Rbb^{K_i}$,
a one-hot encoding of an assignment index at $k = k_i$,
producing $\r = [\r_1, .., \r_I]$.

Seen differently, the indices $(k_1, .., k_I)$
form a discrete product-quantized (PQ) code for $\z \in \Rbb^{DI}$.
PQ codes have been used with great success for indexing in computer vision \cite{jegou2011product}.
It decomposes the vector space $\Rbb^{DI}$ into a Cartesian
product of low dimensional subspaces $\Rbb^D$, and quantizes each
subspace separately, to represent a vector by its subspace quantization indices.

We let $\r_i$ be drawn from a multinomial distribution that is parameterized by $\bpi_i$,
to yield the joint prior distribution % for $p(\z, \r | \bxi)$,
\begin{align}
& p(\r | \bxi) \, p(\z | \r, \bxi) \nonumber \\
& \qquad = \prod_i \prod_k \pi_{ik}^{r_{ik}} \, \Ncal(\z_i ; \bmu_{ik}, \diag(\balpha_{ik})^{-1})^{r_{ik}} . \label{eq:prior}
\end{align}
Our hope is that $(k_1, .., k_I)$ provides a factorial representation of $\x$,
with the ability to choose and combine discrete concepts $k_i$, then generate $\z_i$'s from those discrete
latent variables, and finally to model $\x$ from the concatenation of the $\z_i$'s.

%%%%%%%%%%%%%%%%%%%%%%%%%%%%%%%
\subsection{Hierarchical Model}
%%%%%%%%%%%%%%%%%%%%%%%%%%%%%%%

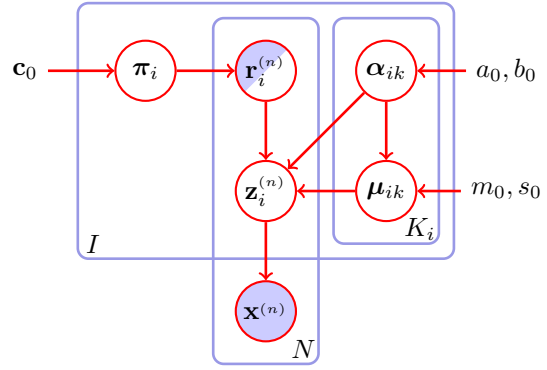
\begin{figure}[t]
\centering
\scalebox{1.0}{
\begin{tikzpicture}[->, line width=1pt, node distance=1.6cm]
% Plate over I
\draw[rounded corners, draw=LightBlue] (-0.9, 0.9) rectangle (4.1, -2.5) {};
\node[draw=none, fill=none] at (-0.7, -2.3) {$I$};
% Plate over K
\draw[rounded corners, draw=LightBlue] (2.5, 0.7) rectangle (3.9, -2.3) {};
\node[draw=none, fill=none] at (3.65, -2.1) {$K_i$};
% Plate over N
\draw[rounded corners, draw=LightBlue] (0.9, 0.7) rectangle (2.3, -3.9) {};
\node[draw=none, fill=none] at (2.1, -3.7) {$N$};
\node[randomv] (pi) {$\bpi_i$};
\node[draw=none, fill=none, left of = pi] (c) {$\c_0$};
\fill[obs, right of = pi] (0,0) -- (45:0.4) arc (45:225:0.4) -- cycle;
% \fill[randomv, obs, right of = pi] (0,0) arc (45:215:1); %  (45:215:1) -- cycle;
\node[randomv, right of = pi] (r) {$\r_i^{\datapoint}$};
\node[randomv, below of = r] (z) {$\z_i^{\datapoint}$};
\node[randomv, obs, below of = z] (x) {$\x^{\datapoint}$};
\node[randomv, right of = r] (alpha) {$\balpha_{ik}$};
\node[randomv, right of = z] (mu) {$\bmu_{ik}$};
\node[draw=none, fill=none, right of = alpha] (ab) {$a_0, b_0$};
\node[draw=none, fill=none, right of = mu] (ms) {$m_0, s_0$};
%
%\node[rnn, right of = gena1t] (x1) {$x^1$};
%
%\node[draw=none, fill=none, below of = x1] (blank2) {};
%\node[rnn, below of = blank2] (x2) {$x^2$};
%\node[randomv, left of = x2] (gena2t) {$a_t^2$};
%\node[randomv, obs, below of = x2] (genvt) {$\x^{\datapoint}$};
%%
\draw[line width=1pt, draw=red](c)--(pi);
\draw[line width=1pt, draw=red](pi)--(r);
\draw[line width=1pt, draw=red](r)--(z);
\draw[line width=1pt, draw=red](z)--(x);
\draw[line width=1pt, draw=red](mu)--(z);
\draw[line width=1pt, draw=red](alpha)--(z);
\draw[line width=1pt, draw=red](alpha)--(mu);
\draw[line width=1pt, draw=red](ab)--(alpha);
\draw[line width=1pt, draw=red](ms)--(mu);
\end{tikzpicture}}
\caption{
The joint density in \eqref{eq:joint}, shown as a probabilistic graphical model.
In the semi-supervised setting, some of the $\r_i^{\datapoint}$ random variables are also observed; see Section \ref{sec:semi-supervised}.
}
\label{fig:generative}
\end{figure}

In \eqref{eq:prior}, the prior is written as $p(\r, \z | \bxi)$, where $\bxi$ are random variables \cite{johnson2016svae}.
As random variables, the hyperprior $p(\bxi)$
is judiciously chosen from a conjugate exponential family of distributions, and
the rich history of variational Bayes from the 1990's and 2000's is drawn on to
additionally
infer the posterior distribution of $\bxi$ \cite{johnson2016svae}.
An alternative, not pursued here, would have been to treat $\bxi$ as model parameters, writing the prior as $p_{\bxi}(\r, \z)$ \cite{fraccaro17disentangled}.

With $\bxi$ in \eqref{eq:prior} being random variables (that are
integrated out when any marginal distributions are determined), they are \emph{a priori} modeled with a hyperprior $p(\bxi)$.
In particular, each Gaussian mean and precision is independently governed by a
conjugate Normal-Gamma hyperprior,
$p(\mu_{ikd}, \alpha_{ikd})
= \Ncal(\mu_{ikd} ; m_0, (s_0 \alpha_{ikd})^{-1}) \, \Gcal(\alpha_{ikd} ; a_0, b_0) 
$, for $d = 1, .., D$.
Our implementation uses $m_0 = 0$, $s_0 = 1$, and $a_0 = b_0 = 0.01$.
The choice of a conjugate hyperprior
becomes apparent when $\bxi$ is inferred from data using stochastic variational inference \cite{hoffman2013stochastic,paquet2013oneclass}.
For a similar reason, each mixing weights vector $\bpi_i$ is drawn from a Dirichlet distribution $p(\bpi_i) = \Dcal(\bpi_i ; \c_0)$
with pseudo-counts $\c_0 = \1$.
The joint density over the observed random variables $\X$ and hidden random variables
$\Z = \{ \z^{\datapoint} \}_{n=1}^N$, $\R = \{ \r^{\datapoint} \}_{n=1}^N$ and $\bxi$ is
\begin{align}
p(\X, \Z, \R, \bxi) 
= & \prod_n p_{\btheta}(\x^{\datapoint} | \z^{\datapoint}) \,
p(\z^{\datapoint} | \r^{\datapoint}, \bxi) \,
p(\r^{\datapoint} | \bxi) \nonumber \\
& \cdot \prod_i p(\bpi_i) \prod_k p(\bmu_{ik}, \balpha_{ik}) \ , \label{eq:joint}
\end{align}
and is illustrated as a probabilistic graphical model in Figure \ref{fig:generative}.
Equation \eqref{eq:joint} contains local factors and global random variables,
and the posterior approximation in Section \ref{sec:posterior-approximation}
will contain local and global factors.

The model's prior is related to an independent component analysis (ICA) prior on $\z$, after which a non-linear mixing function is applied to yield $\x$.
It is also worth noting that $I=1$ yields \citet{johnson2016svae}'s
Gaussian mixture model (GMM) structured variational auto-encoder (SVAE),
and we generalise here from a single GMM to ICA-style component-wise mixtures.
When $\bxi$ is treated as parameters (and not random variables) in $p_{\bxi}(\r, \z)$, and $I$ set to one, we obtain
\citet{jiang2016variational}'s variational deep embedding (VaDE) model.
Further connections to related work are made in Section \ref{sec:related}.

%%%%%%%%%%%%%%%%%%%%%%%%%%%%%%%%%%%%
\subsection{Posterior Approximation}
\label{sec:posterior-approximation}
%%%%%%%%%%%%%%%%%%%%%%%%%%%%%%%%%%%%

We approximate the posterior with a fully factorized distribution
\begin{align} 
p(\Z, \R, \bxi \, | \, \X)
\approx &
\underbrace{ {\textstyle\prod}_n \, q_{\bphi}(\z^{\datapoint} | \x^{\datapoint}) }_{q_{\bphi}(\Z | \X)} \,
\underbrace{ {\textstyle\prod}_n \, q_{\bgamma}(\r^{\datapoint}) }_{q_{\bgamma}(\R)} \nonumber \\
& \cdot \underbrace{ {\textstyle\prod}_i \, q_{\blambda}(\bpi_i) \,
{\textstyle\prod}_k \, q_{\blambda}(\bmu_{ik}, \balpha_{ik}) }_{q_{\blambda}(\bxi)} \ .
\label{eq:approx}
\end{align}
An ``encoding network'' or ``inference network''
$q_{\bphi}(\z | \x)
= \Ncal( \z \, ; \, \bmu_{\bphi}(\x), \, \diag \bsigma_{\bphi}^2(\x) )$
amortizes inference for $\z$, and
maps $\x^{\datapoint}$ to a Gaussian distribution on $\z^{\datapoint}$;
both $\bmu_{\bphi}(\x)$ and $\bsigma_{\bphi}^2(\x)$ are deep neural networks \cite{kingma2013auto}.
We approximate $q_{\bgamma}(\R)$ and $q_{\blambda}(\bxi)$ with a ``structured VAE'' approach \cite{johnson2016svae}.

There is a local approximation $q_{\bgamma}(\r_i^{\datapoint}) = \prod_k [\gamma_{ik}^{\datapoint}]^{r_{ik}^{\datapoint}}$
for every data point, yielding a ``responsibility'' for block $i$ through $\sum_k \gamma_{ik} = 1$.
% The soft responsibilities for the full latent vector is
Furthermore,
$q_{\bgamma}(\r^{\datapoint}) = \prod_i q_{\bgamma}(\r_i^{\datapoint})$.
Define $\bgamma = \{ \{ \bgamma_i^{\datapoint} \}_{i=1}^I \}_{n=1}^N$
as parameters of $q(\R)$.
Maximization for the local factor $q_{\bgamma}(\r^{\datapoint})$ is semi-amortized;
we'll see in \eqref{eq:resp} that it can be obtained
as an analytic, closed form expression.
In \eqref{eq:resp} it is a softmax over cluster assignments,
and could be interpreted as another ``encoding network'',
this time for the
one-hot discrete representations $\r_i$.\footnote{
Here, there is room for splitting hairs.
We opt for uncluttered notation $q_{\bgamma}(\r_i^{\datapoint})$, following the traditional
convention in the variational Bayes (VB) literature.
According to \eqref{eq:resp}, the ELBO is locally maximized at an expression
for
$q_{\bgamma}(\r_i^{\datapoint})$ that depends on $\bphi$, $\blambda$
and $\x^{\datapoint}$. One might also make the ``amortized''
nature of these local factors clearer by writing the factors as
$q_{\bphi \blambda}(\r_i^{\datapoint} | \x^{\datapoint})$.}

To express $q_{\blambda}(\bxi)$, we parameterize each of the factors
\begin{align*}
& q_{\blambda}(\bmu_{ik}, \balpha_{ik}) = \textstyle{\prod}_d \, q(\mu_{ikd}, \alpha_{ikd}) \\
& = \textstyle{\prod}_d \,
\Ncal(\mu_{ikd} ; m_{ikd}, (s_{ikd} \alpha_{ikd})^{-1}) \, \Gcal(\alpha_{ikd} ; a_{ikd}, b_{ikd})
\end{align*}
as a product of Normal-Gamma distributions,
in the same form as the prior.
Lastly, let $q_{\blambda}(\bpi_i)$ be a Dirichlet distribution parameterized by
its pseudo-counts vector $\c_i$.

We've stated the factors in terms of their \emph{mean parameters}
$\{ \{ \m_{ik}, \s_{ik}, \a_{ik}, \b_{ik} \}_{k=1}^{K_i}, \c_i \}_{i=1}^I$
here because we
require the mean parameters in \eqref{eq:Elogp}.
However, instead of doing gradient descent in the mean parameterization, we will view 
$q_{\blambda}(\bxi)$ through the lens of its \emph{natural parameters} $\blambda$.
This parameterization has pleasing properties, as it allows gradients to be conveniently expressed as natural gradients;
a gradient step of length one gives an update to a local minimum (for a batch) \cite{paquet2015convergence}.
This is also the framework required for stochastic variational inference (SVI) \cite{hoffman2013stochastic}.
Here the exponential family representation of $q_{\blambda}(\bxi)$ is minimal and
there exists a one-to-one mapping between the mean parameters and natural parameters $\blambda$ \cite{wainwright2008graphical};
we state the mapping in Appendix \ref{app:natural}.

%%%%%%%%%%%%%%%%%%%%%%%%%%%%%%%%%%%
\subsection{Inference and Learning}
\label{sec:inference}
%%%%%%%%%%%%%%%%%%%%%%%%%%%%%%%%%%%

The posterior approximation depends on $\btheta$, $\bphi$, $\blambda$, and $\bgamma$.
They will be found by maximizing a variational lower bound to $\log p_{\btheta}(\X) \ge \Ebb_{q}[\log p_{\btheta}(\X, \Z, \R, \bxi) - \log q_{\bphi}(\Z | \X) \, q_{\bgamma}(\R) \, q_{\blambda}(\bxi) ]  = \Lcal(\btheta, \bphi, \blambda, \bgamma)$, also referred to as the
``evidence lower bound'' (ELBO):
\begin{subequations}
\begin{align}
& \Lcal(\btheta, \bphi, \blambda, \bgamma) \nonumber \\
& = \sum_n
\Bigg\{
\Ebb_{q_{\bphi}(\z^{\datapoint} | \x^{\datapoint})} \left[ \log p_{\btheta}(\x^{\datapoint} | \z^{\datapoint}) \right]
\label{eq:elbo:a} \\
& \qquad - \sum_{i,k}
\underbrace{ \Ebb_{q_{\bgamma}(\r_i^{\datapoint})} [r_{ik}^{\datapoint}] }_{
\gamma_{ik}^{\datapoint} \textrm{from \eqref{eq:resp} at max}
} \, \Ebb_{q_{\blambda}(\bmu_{ik}, \balpha_{ik})} \Big[ \cdots \nonumber \\
& \qquad \quad \cdots \KL \Big(
q_{\bphi}(\z_i^{\datapoint} | \x^{\datapoint}) \, \Big\| \,
p(\z_i^{\datapoint} | \bmu_{ik}, \balpha_{ik}) \Big) \Big]
\label{eq:elbo:b} \\
%%%
& \qquad - \sum_i \Ebb_{q_{\blambda}(\bpi_i)} \left[ \KL \Big( q_{\bgamma}(\r_i^{\datapoint}) \, \Big\| \, p(\r_i^{\datapoint} | \bpi_i) \Big) \right]
\label{eq:elbo:c} \\
& \qquad - \frac{1}{N} \sum_{i,k} \KL \Big( q_{\blambda}(\bmu_{ik}, \balpha_{ik}) \, \Big\| \, p(\bmu_{ik}, \balpha_{ik}) \Big) \label{eq:elbo:d} \\
& \qquad - \frac{1}{N} \sum_i \KL \Big( q_{\blambda}(\bpi_i) \, \Big\| \, p(\bpi_i) \Big) \Bigg\} \ . \label{eq:elbo:e}
\end{align}
\end{subequations}
The full derivation is provided in Appendix \ref{app:elbo}.
The contribution of the hyperpriors is equally divided over all data points through $\sum_n \frac{1}{N}$ in lines \eqref{eq:elbo:d} and \eqref{eq:elbo:e}, so that the outer sum is over $n$.
This aids stochastic gradient descent, and appropriately weighs $q_{\blambda}(\bxi)$'s gradients as part of mini-batches of $\X$.
The objective in \eqref{eq:elbo:a} to \eqref{eq:elbo:e} forms a
structured VAE \cite{johnson2016svae}.

There are four Kullback-Leibler ($\KL$) divergences in $\Lcal$.
For each data point and factor $i$, \eqref{eq:elbo:b} contains 
the KL divergence between
$q_{\bphi}(\z_i | \x)$ and $p(\z_i | \bmu_{ik}, \balpha_{ik})$.
This is the familiar local KL divergence between the encoder and the prior.
Because $\bxi$ is inferred, the KL is averaged over the (approximate) posterior uncertainty $q_{\blambda}(\bmu_{ik}, \balpha_{ik})$. 
With the allocation of $\x$ to clusters $(k_1, .., k_I)$ also being inferred,
the smoothed KL is further weighted in a convex combination over $\Ebb_{q} [r_{ik}]$.
A second local KL cost is incurred in line \eqref{eq:elbo:c}, where discrete random variables $\r$ encode information about $\x$.
Finally, lines \eqref{eq:elbo:d} and \eqref{eq:elbo:e} penalize
the global posterior approximation $q_{\blambda}(\bxi)$ for moving from the prior $p(\bxi)$.
The KL divergence between two Normal-Gamma distributions is given in Appendix \ref{app:normal-gamma-kl}.

Objective  $\Lcal$ can be optimized using three different techniques,
spanning the last three decades of progress in variational inference.
This combination has already been employed by a number of authors recently \cite{johnson2016svae, lin2018variational}.
Given a mini-batch $\Bcal$ of data points $n$ at iteration $\tau$, the following parameter updates are followed:
\begin{enumerate}
\item $\Lcal$ is locally maximized with respect to $\bgamma_i^{\datapoint}$ with an analytic, closed form expression.
In an Expectation-Maximization (EM) algorithm, this is the variational Bayes E-step \cite{attias1999inferring,waterhouse1996bayesian}.
\item Given a batch's $\{ \bgamma_i^{\datapoint} \}$,
a stochastic gradient descent step updates $\btheta$ and $\bphi$,
for example using
\citet{kingma2014adam}'s optimizer.
When line \eqref{eq:elbo:b} is expressed analytically using \eqref{eq:Elogp},
the reparameterization trick can be readily used \cite{kingma2013auto}.
\item After recomputing $\{ \bgamma_i^{\datapoint} \}$
at their new local optima using the updated
$\btheta$ and $\bphi$,
a SVI natural gradient step is taken on
$\blambda$, using a decreasing Robbins-Monro step size \cite{hoffman2013stochastic}.
The updates on $\btheta$, $\bphi$  and $\blambda$ together
form a stochastic M-step.
\end{enumerate}
In practice, we initialize $\btheta$ and $\bphi$ by first running a number of gradient updates, using an ELBO with a $\Ncal(\z ; \0, \I)$ prior, possibly with an annealed KL term. That gives an initial encoder $p_{\bphi}(\z | \x)$.
Parameters $\blambda$ are then initialized by repeating steps 1 and 3, to seed
a basic factorized clustering of the encoded $\z$'s.
We consider steps 1--3 separately below:

%%%%%%%%%%%%%%%%%%%%%%%%%%%%%%%%%%%%%%%%%%%%%%%%%%%%%%%%%%%%%%
\subsubsection{Locally maximizing over $\bgamma^{\datapoint}$}
%%%%%%%%%%%%%%%%%%%%%%%%%%%%%%%%%%%%%%%%%%%%%%%%%%%%%%%%%%%%%%
The local maximum of $\Lcal$ over $q_{\bgamma}(\r_{i}^{\datapoint})$ is the softmax
\begin{align}
\gamma_{ik} & = 
\widetilde{\gamma}_{ik} \Big/ \textstyle{\sum_{k'}} \widetilde{\gamma}_{ik'} \label{eq:resp} 
\\
\widetilde{\gamma}_{ik} & =
\exp \{ \, \Ebb_{q_{\bphi}(\z_i | \x) \, q_{\blambda}(\bmu_{ik}, \balpha_{ik})} [\log p(\z_i | \bmu_{ik}, \balpha_{ik})] \nonumber \\
& \qquad\qquad
\qquad\qquad
\qquad + \Ebb_{q_{\blambda}(\bpi_i)}[\log \pi_{ik}] \, \} \nonumber
\end{align}
(dropping superscripts $n$).
If the encoding network $q_{\bphi}(\z | \x)$ yields a Gaussian distribution,
we can determine the above expectation required for \eqref{eq:resp} analytically:
\begin{align}
& \Ebb_{q_{\bphi}(\z_i | \x) \, q_{\blambda}(\bmu_{ik}, \balpha_{ik})} [\log \Ncal(\z_i ; \bmu_{ik}, \diag(\balpha_{ik})^{-1})] \nonumber \\
& = 
\frac{1}{2} \sum_d \Bigg( \psi(a_{ikd}) - \log b_{ikd} -  \log (2 \pi)
- \frac{1}{s_{ikd}}
\nonumber \\
& \qquad - \frac{a_{ikd}}{b_{ikd}} \Big( (\bmu_{\bphi}(\x)_{d'} - m_{ikd})^2 + \bsigma_{\bphi}^2(\x)_{d'} \Big)
\Bigg) \label{eq:Elogp}
\end{align}
where index $d' = d + (i - 1)D$ is the index into the matching sub-vector in $\z$,
and $\psi(\cdot)$ indicates the digamma function.
Furthermore, $\Ebb_{q_{\blambda}(\bpi_i)}[\log \pi_{ik}] = \psi(c_{ik}) - \psi(\sum_k c_{ik})$.
The expression in \eqref{eq:Elogp} appears in line \eqref{eq:elbo:b},
so that the KL divergence is expressed exactly
when stochastic gradients over $\btheta$ and $\bphi$ are computed.

Aside from showing that the maximum of $\Lcal$ with respect to $q_{\bgamma}(\r_{i}^{\datapoint})$
is analytically tractable, \eqref{eq:resp} is a softmax classification into expected cluster allocations.
Semi-supervised labels could be used to guide $q_{\bphi}(\z_i | \x)$---since it appears in analytic form in \eqref{eq:Elogp}---towards
encodings for which the expected cluster allocations correspond
to labelled properties or attributes;
we turn to this theme in \eqref{eq:semi-sup} in Section \ref{sec:semi-supervised}.

%%%%%%%%%%%%%%%%%%%%%%%%%%%%%%%%%%%%%%%%%%%%%%%%%%%%%%%%%%
\subsubsection{Stochastic gradients of $\btheta$, $\bphi$}
%%%%%%%%%%%%%%%%%%%%%%%%%%%%%%%%%%%%%%%%%%%%%%%%%%%%%%%%%%

Lines \eqref{eq:elbo:a} and \eqref{eq:elbo:b} depend on $\btheta$ and $\bphi$,
and the ``reparameterization trick'' can be used with a stochastic gradient optimizer \cite{kingma2014adam,kingma2013auto}.

%%%%%%%%%%%%%%%%%%%%%%%%%%%%%%%%%%%%%%%%%%%%%%%%%%%%%%%%%%
\subsubsection{Conjugate, natural gradients of $\blambda$}
\label{sec:conjugate}
%%%%%%%%%%%%%%%%%%%%%%%%%%%%%%%%%%%%%%%%%%%%%%%%%%%%%%%%%%

We provide only an sketch of natural gradients and their use in SVI here, and provide the
derivations and detailed expressions of the gradients in
Appendices \ref{app:natural} and \ref{app:natural-gradients}.

At iteration $\tau$, for mini-batch $\Bcal$ of data points,
$\Lcal$ is analytically minimized with respect to
$\blambda$ to yield $\blambda^*$.
If $\blambda^{(\tau)}$ denotes the natural parameters at iteration $\tau$,
then the natural gradient is
$\widehat{\nabla}_{\blambda^{(\tau)}} \Lcal(\blambda) = \blambda^* - \blambda^{(\tau)}$ \cite{paquet2015convergence}.
We update
$\blambda^{\tau + 1} \leftarrow
\blambda^{\tau} + \rho_{\tau} \widehat{\nabla}_{\blambda^{\tau}} \Lcal(\blambda)$,
which is equivalent to updating it to the convex combination between
the currently tracked minimizer $\blambda^{(\tau)}$ and the
batch's minimizer $\blambda^*$,
\begin{equation} \label{eq:lambda-update}
\blambda^{(\tau + 1)} \leftarrow (1 - \rho_{\tau}) \blambda^{(\tau)} + \rho_{\tau} \blambda^* \ .
\end{equation}
This parameter update is a rewritten from of updating 
$\blambda^{(\tau)}$ with the natural gradient $\widehat{\nabla}_{\blambda^{(\tau)}} \Lcal(\blambda)$
that is rescaled by $\rho_{\tau}$ \cite{hoffman2013stochastic}.
The step sizes should satisfy $\sum_{\tau = 1}^{\infty} \rho_{\tau} = \infty$ and $\sum_{\tau = 1}^{\infty} \rho_{\tau}^2 < \infty$.
In our results $\rho_{\tau} = (\tau_0 + \tau)^{-\kappa}$ was used,
with forgetting rate $\kappa \in (\frac{1}{2}, 1]$
and delay $\tau_0 \ge 0$.
The updated \emph{mean parameters} of $q_{\blambda}(\bxi)$ are obtained from the updated natural parameters $\blambda^{(\tau + 1)}$.

%%%%%%%%%%%%%%%%%%%%%%%%%%%%%%%%%%%%%
\subsection{Semi-supervised Learning}
\label{sec:semi-supervised}
%%%%%%%%%%%%%%%%%%%%%%%%%%%%%%%%%%%%%

For some data points $n \in \Omega$, we have access to labels $\y^{\datapoint}$.
With appropriate preprocessing, we let $\y_i^{\datapoint}$ correspond to a one-hot encoding for property $i$, and let $i \in \Pi(n)$ denote the observed properties for data point $n$.
Where present, we treat the labels as \emph{observed random variables}
$\r_i^{\datapoint} = \y_i^{\datapoint}$ in the probabilistic graphical model in
\eqref{eq:joint} and Figure \ref{fig:generative}.
The posterior approximation in \eqref{eq:approx},
which is used in \eqref{eq:elbo:b} and \eqref{eq:elbo:c},
hence includes pre-set delta functions
$q_{\bgamma}(\r_i^{\datapoint}) = \delta(\r_i^{\datapoint} - \y_i^{\datapoint})$.
Certain marginals in the posterior approximation are thus clamped.

Merely clamping marginals may not be sufficient
for ensuring that $\z_1$ encodes, for example, latent
\texttt{male} or \texttt{female} properties in Figure
\ref{fig:faces-and-glasses}.
The encoder $q_{\bphi}(\z | \x)$
might still distribute these properties through \emph{all} of $\z$,
and decoder $p_{\btheta}(\x | \z)$ find parameters that recover $\x$ from that $\z$.
We would like the encoder to emit
two linearly separable clusters for block $\z_1$ for \texttt{male} and \texttt{female} inputs, and define
the expected cluster allocation
$g_{ik}^{\datapoint} = \widetilde{\gamma}_{ik}^{\datapoint} / \sum_{k'} \widetilde{\gamma}_{ik'}^{\datapoint}$
like \eqref{eq:resp}.
To encourage a distributed representation of
\texttt{male} and \texttt{female} primarily in $\z_1$,
we incorporate a cross entropy classification loss
to $\Lcal(\btheta, \bphi, \blambda, \bgamma)$,
\begin{align}
\Fcal(\btheta, \bphi, \blambda, \bgamma) = \Lcal + \Delta \sum_{
n \in \Omega
% \substack{n \in \Omega \\ i \in \Pi(n)}
} \sum_{i \in \Pi(n)}
\sum_k y_{ik}^{\datapoint} \log g_{ik}^{\datapoint} , \label{eq:semi-sup}
\end{align}
with a tunable scalar knob $\Delta \ge 0$.

On closer inspection, the ``logits'' of the $g_{ik}^{\datapoint}$'s
are differentiable functions of the encoder,
using the analytic expression in
\eqref{eq:Elogp}.
For the parameters of the subset of $q_{\blambda}(\bpi_i)$ and 
$q_{\blambda}(\bmu_{ik}, \balpha_{ik})$ factors that influence the
semi-supervised loss term in \eqref{eq:semi-sup},
the closed-form natural gradient updates of Section \ref{sec:conjugate} are no longer possible.
When $\Delta > 0$, the M-step can employ first-order gradients in the mean parameterization of those factors.

As a final thought, we are free to set $\Delta = 0$ once we are satisfied that 
$q_{\bphi}(\z | \x)$ primarily represents a latent \texttt{male} and \texttt{female} signal in $\z_1$, for instance.
In that case we are left with a probabilistic graphical model in which some $\r_i^{\datapoint}$ are observed, hence
still encouraging a separation of representation of named properties.

%%%%%%%%%%%%%%%%%%%%%
\subsection{Sampling}
\label{sec:sampling}
%%%%%%%%%%%%%%%%%%%%

We wish to sample from the marginal distribution of $\x$, conditional on all observed data $\X$. It is an intractable average
over the posterior $p(\bxi | \X)$,
\begin{align}
& p_{\btheta}(\x | \X)
= \sum_{\r} \int p_{\btheta}(\x, \z, \r, \bxi | \X) \, \drm \z \, \drm \bxi \nonumber \\
& \quad = \sum_{\r} \int p_{\btheta}(\x | \z) \, p(\z | \r, \bxi) \, p(\r | \bxi) \, p(\bxi | \X) \, \drm \z \, \drm \bxi \ , \label{eq:conditional}
\end{align}
which we approximate by substituting $p(\bxi | \X)$ with $q(\bxi)$.
Call the distribution resulting from this substitution $p_{\Qcal}(\x) \approx p_{\btheta}(\x | \X)$.
To sample from $p_{\Qcal}(\x)$,
one-hot cluster indices are first sampled with $\bpi_i \sim q_{\blambda}(\bpi_i)$ and
$\r_i \sim \Ccal(\bpi_{i})$,
a categorical distribution with $\bpi_i$ as its event probabilities.
Then, let $k_i$ be the non-zero index of $\r_i$,
and sample $\z_i \sim q_{\blambda}(\z_i | k_i) = \int p(\z_i | \bmu_{ik}, \balpha_{ik})
q_{\blambda}(\bmu_{ik}, \balpha_{ik}) \, \drm \bmu_{ik} \balpha_{ik}$.
Each $z_{d'}$ is drawn from a heavy-tailed Student-t distribution
with $2 a_{ikd}$ degrees of freedom
\begin{equation} \label{eq:z-t-dist}
q_{\blambda}(z_{d'} | k_i) = \Tcal\left(z_{d'} ; m_{ikd}, \frac{s_{ikd} + 1}{s_{ikd}} \frac{b_{ikd}}{a_{ikd}}, 2 a_{ikd} \right) \ ,
\end{equation}
where $d' = d + (i - 1)D$ indexes block $i$'s entry $d$.
Finally, sample $\x \sim p_{\btheta}(\x | \z)$.

\subsection{Predictive Log Likelihood}

The predictive likelihood of a new data point $\x$ is estimated with a lower bound to $p_{\Qcal}(\x)$,
%%%
which decomposes with two terms containing KL divergences,
\begin{align}
& \log p_{\Qcal}(\x) 
\ge \Ebb_{q_{\bphi}(\z | \x)} \left[ \log p_{\btheta}(\x | \z) \right] \nonumber \\
& -
\underbrace{ \Ebb_{q_{\blambda}(\bmu, \balpha)} \left[ \sum_{i,k} \Ebb_q[ r_{ik} ] \, 
\KL \Big( q_{\bphi}(\z_i | \x) \, \Big\| \, p(\z_i | \bmu_{ik}, \balpha_{ik}) \Big) \right] }_{\KL[\z]}
\nonumber \\
& -
\underbrace{ \Ebb_{q_{\blambda}(\bpi)} \left[ \sum_i \KL \Big( q_{\bgamma}(\r_i) \, \Big\| \, p(\r_i | \bpi_i) \Big)
\right] }_{\KL[\r]}
\ . \label{eq:test-elbo}
\end{align}
The bound in \eqref{eq:test-elbo} is first maximized over $q(\r_i)$, using \eqref{eq:resp}, before being evaluated.
In our results, we consider the tradeoff between encoding in $\z$ and in $\r$, by evaluating the
two terms $\KL[\z]$ and $\KL[\r]$.

%!TeX root = main.tex

\section{Related Work}
\label{sec:related}

The factorial representation in this paper is a collection of ``Gaussian mixture model structured VAEs (SVAEs)'',
and our optimization scheme in Section \ref{sec:inference} mirrors that of Algorithm 1 in \cite{johnson2016svae}.
In a larger historical context, the model has roots in variational
approximations for mixtures of factor analyzers \cite{beal2000variational}.
In our work, the prior $p(\z | \r, \bxi)$ is a Gaussian on $\z$, and so
is $q_{\bphi}(\z | \x)$. As the product of two Gaussians yields an unnormalized Gaussian (with closed form normalizer), 
the prior structure can further be incorporated in the
inference network of a VAE \cite{lin2018variational}.
% , building on \cite{johnson2016svae}.

When the conjugate hyperprior is ignored and $\bxi$ treated as parameters,
and a single mixture set at $I = 1$, we recover the model of \cite{jiang2016variational}.
Alternatively, the ``stick-breaking VAE'' uses
a stick-breaking prior on $\bpi$ for $I = 1$, 
and treats all other $\bxi$ as parameters 
\cite{nalisnick2017stick,nalisnick2016approximate}.
The Gaussian mixture VAE \cite{dilokthanakul16}
uses deep networks to transform
a sample from a unit-variance isotropic Gaussian into a
the mean and variance for each mixture component of
a Gaussian mixture model prior on $\z$. Instead of inferring separate means and variances of the mixture
components, they are learned non-linear transformations
of the same random variable.
Another variation on the mixture-theme, the ``variational
mixture of posteriors'' prior writes the mixture
prior as a mixture of inference networks,
conditioned on learnable pseudo-inputs \cite{tomczak2017vae}.
In \cite{graves2018associative}, the notion of a mixture prior is taken to the extreme, storing the approximate posterior distributions of every element in the training set as mixture components,
and using hard $k$-nearest neighbour lookups to determine the cluster assignment for a new $\x$.

The vector-quantized VAE (VQ-VAE) model \cite{vandenoord2017neural}
is similar to the work presented here, with these differences: here, there are three types random variables,
$\r^{\datapoint}$ (discrete) together with $\z^{\datapoint}$ and $\bxi$,
while VQ-VAE consists only of discrete random variable (a form of $\r^{\datapoint}$).
VQ-VAE introduces a ``stop-gradient'', which on the surface resembles an E-step in an EM algorithm
(see Section \ref{sec:inference}).

A growing body of literature exists around obtaining ``untangled'' representations from unsupervised models.
These range from bootstrapping partial supervised data \cite{narayanaswamy2017}, changing the relative contribution of the KL term(s) in the ELBO \cite{higgins2017beta} to
total correlation penalties \cite{chen2018isolating,kim2018disentangling}
to specialized domain-adverserial training on the
latent space \cite{lample2017fader}.
In this work, the only additional penalties that are added to the ELBO appear in the form of the semi-supervised loss term in \eqref{eq:semi-sup}.
% nor are terms artificially scaled.
The subdivision of data into a discrete and somewhat interpretable Cartesian product of clusters is purely the result of a hierarchical mixture model.
The model presented in this paper can be viewed as a VAE with latent
code vector $\z$ with a ``KD encoding'' prior \cite{chen18kway},
learned in a Bayesian probabilistic way.

Instead of letting $\y$ be parent random variables of $\z$, they can be treated as ``side information'' whose prediction helps recover an interpretable common latent variable $\z^*$.
\citet{adel18discovering}'s ``interpretable lens variable model'' constructs an invertible mapping (normalizing flow) between $\z$ and $\z^*$,
and uses it to add an interpretable layer to,
for example, pre-trained  $p_{\btheta}(\x | \z)$'s and $q_{\bphi}(\z | \x)$'s.
Like this work, $|\Omega|$ might be much smaller than $N$.

% !TeX root = main.tex

\section{Experimental results}
\label{sec:results}

We expand the example of Figure \ref{fig:faces-and-glasses},
by illustrate a factorization of CelebA faces into a product of cluster indices $(k_1, .., k_5)$,
the first three of which are interpretable.
Using the binary MNIST and Omniglot data sets,
we empirically show the
trade-off between costs for representing $\x$ as discrete $\r$ and continuous $\z$ latent variables in the predictive log likelihood \eqref{eq:test-elbo}, as 
$K$ varies.

%We show experimentally how the factorial mixture model approximates the density of three training data sets, MNIST, Omniglot, and CelebA.
%Aside from the expressiveness of $p_{\btheta}(\x | \z)$,
%the ELBO, as a marginal likelihood bound, is a function of the dimensionality of the random variables it is evaluated over.
%We show on binarized MNIST and Omniglot that a classic ``Occam hill'' arises as $K$ is increased \cite{attias1999inferring,paquet2009perturbation},
% and substantially improve the state-of-the-art test ELBO for Omniglot.
%On CelebA, we show an interpretable factorization into a Cartesian product of two clusters $(k_1, k_2)$.

% !TeX root = main.tex

\subsection{Binary MNIST and Omniglot}

\begin{figure*}[t]
\centering
\includegraphics[width=0.9\columnwidth]{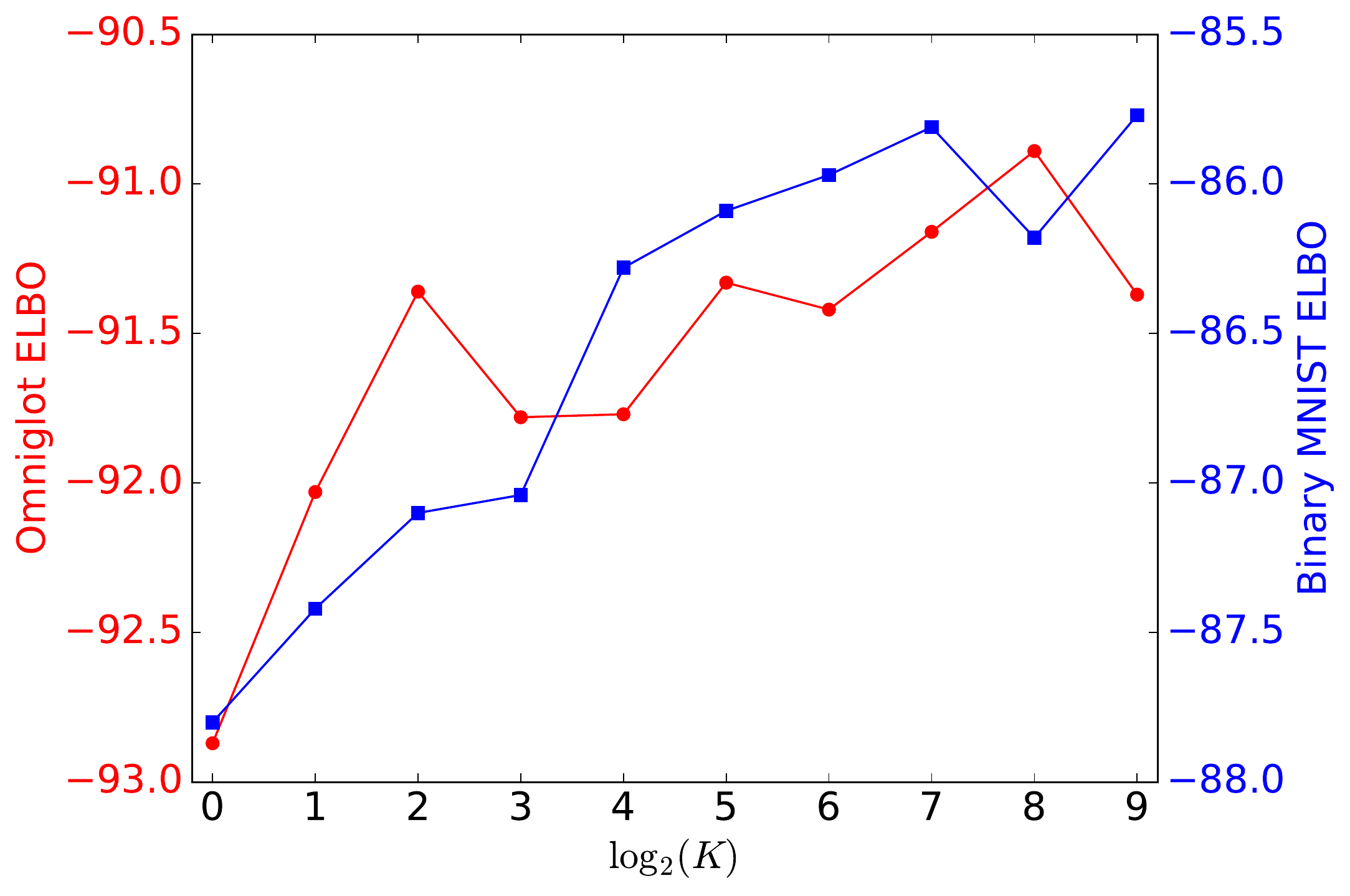} \phantom{ccc}
\includegraphics[width=0.9\columnwidth]{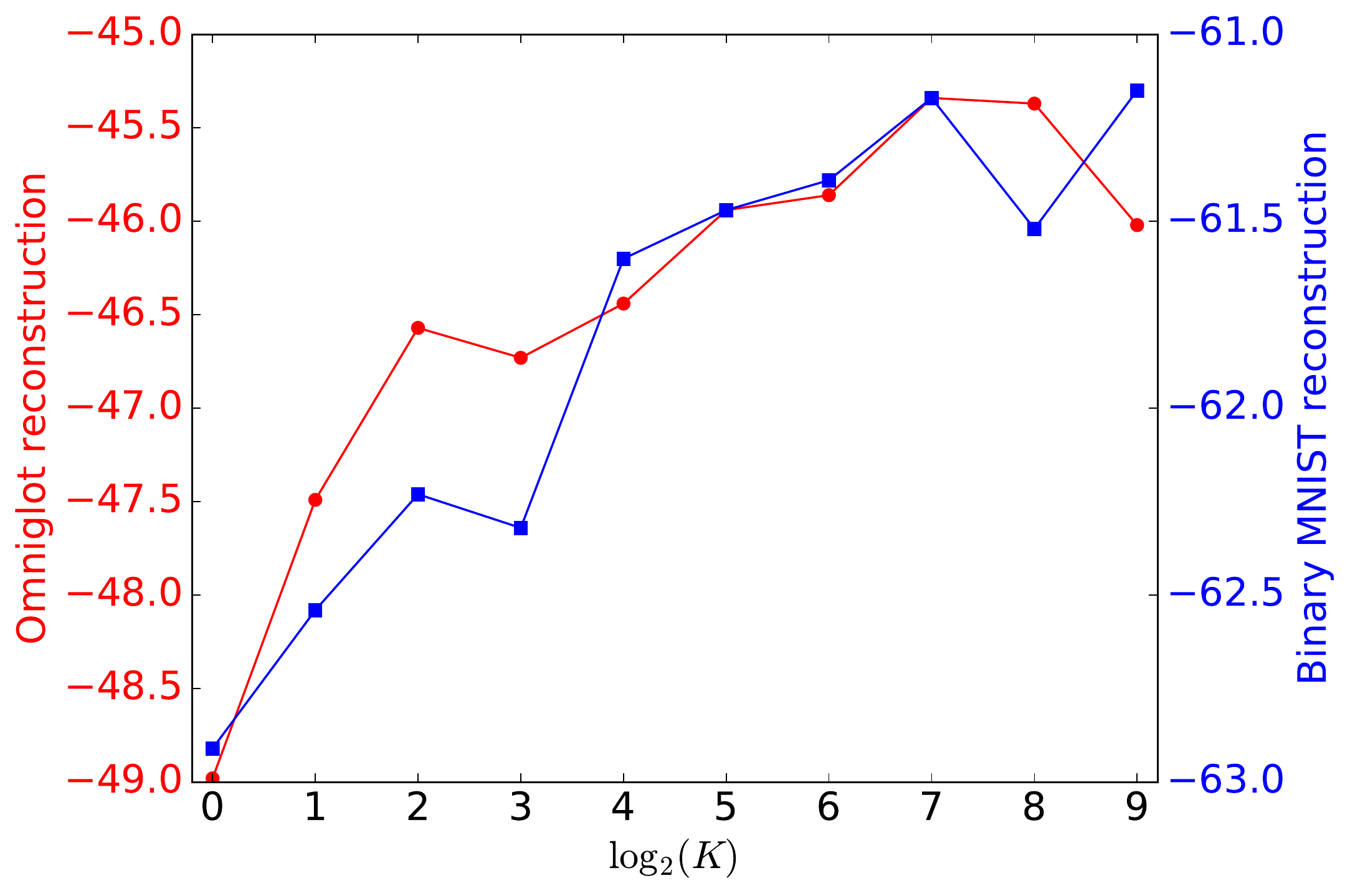} \\
\includegraphics[width=0.9\columnwidth]{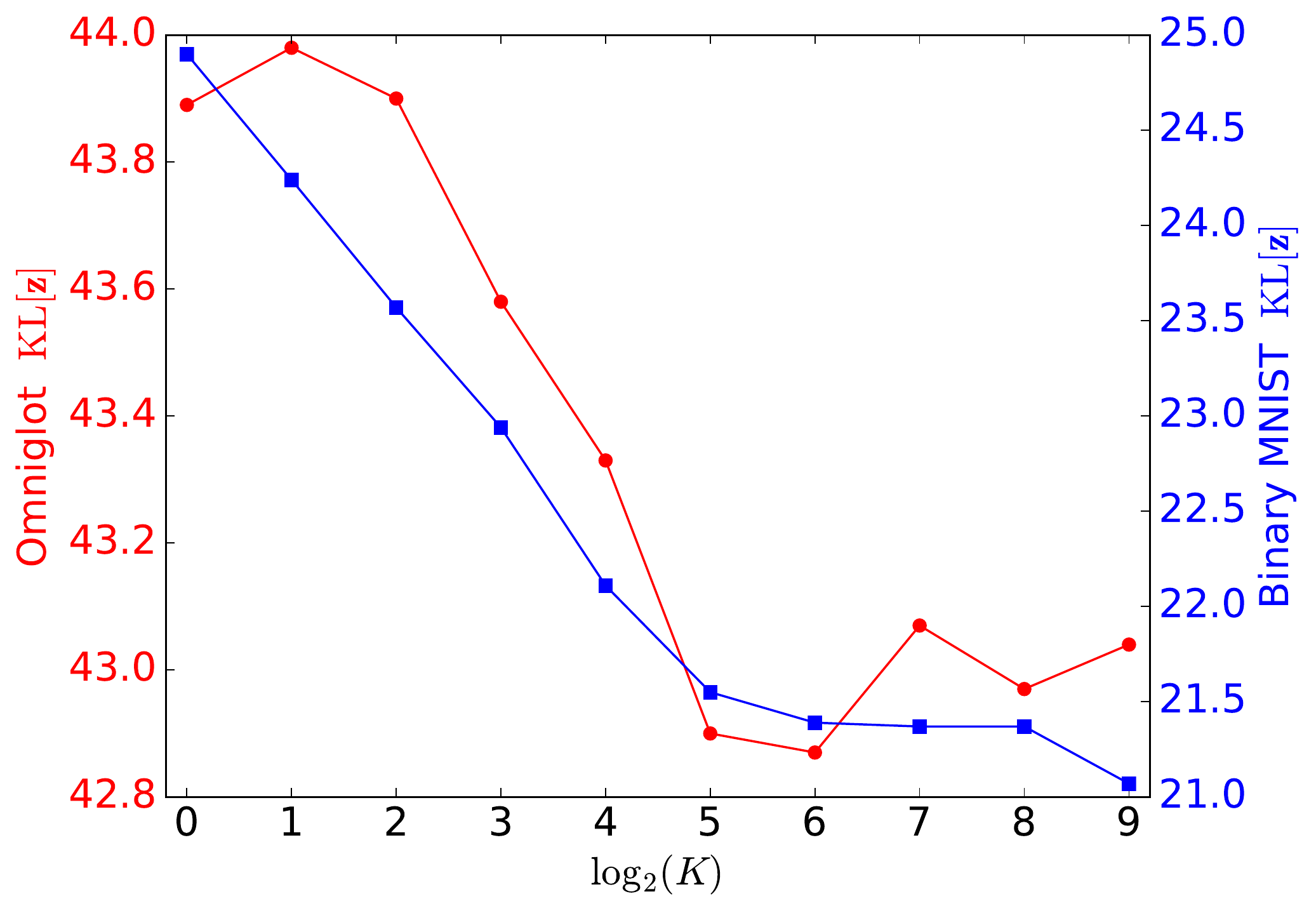} \phantom{ccc}
\includegraphics[width=0.9\columnwidth]{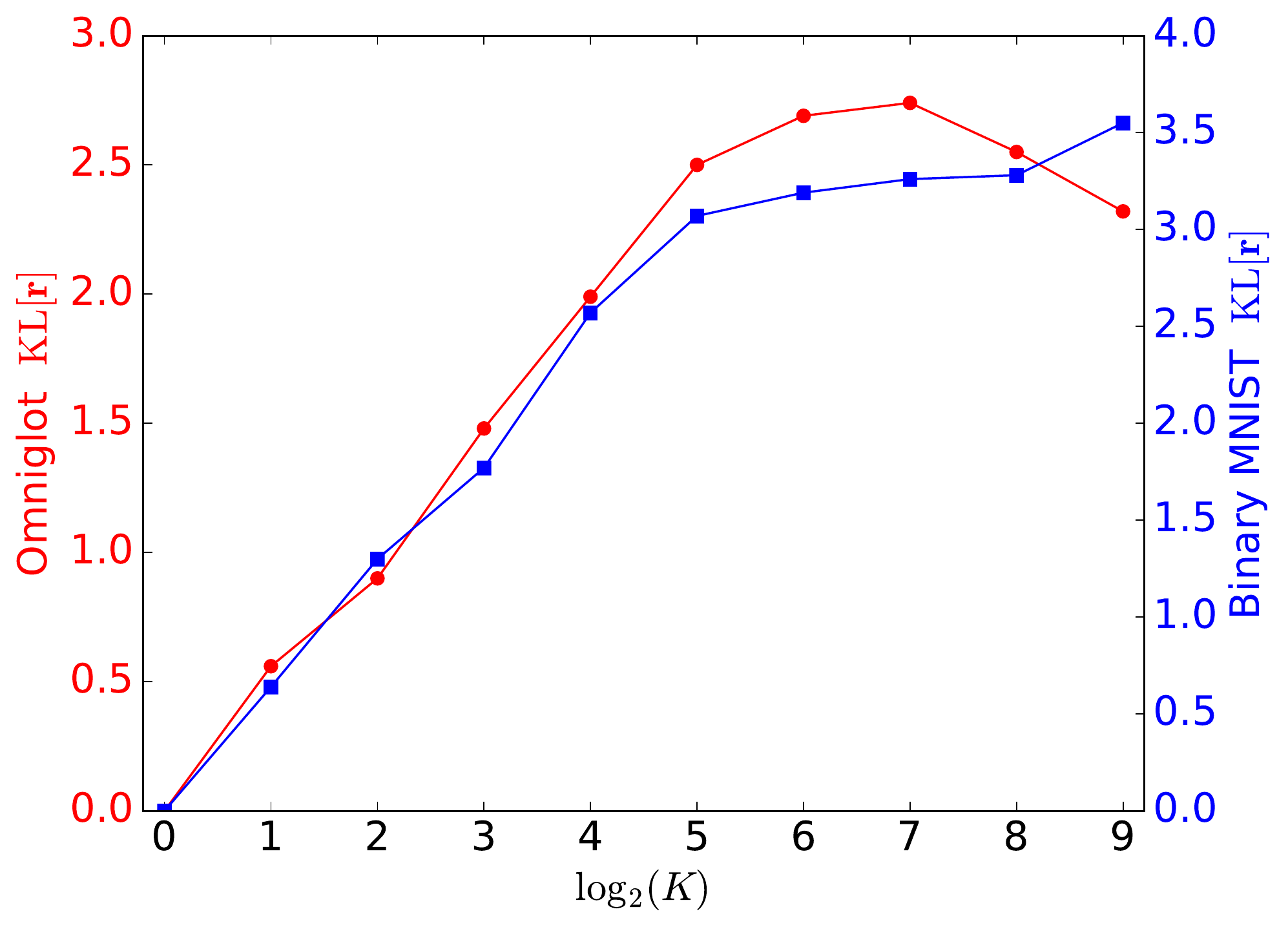}
\caption{The ELBO of \eqref{eq:test-elbo},
averaged over respective tests sets,
as a function of $\log_2 K$.
Generally, the cost $\KL[\z]$ of encoding $\x$ decreases with $K$, at the expense of a higher relative entropy $\KL[\r]$.
More mixture components aid the ``reconstruction term'' $\Ebb_{q_{\bphi}(\z | \x)} [ \log p_{\btheta}(\x | \z)] $, also shown as
averaged over the respective test sets.
% Equation \eqref{eq:test-elbo} does not incorporate
% the global KL cost for $\bxi$, as we see in
% \eqref{eq:elbo:d} and \eqref{eq:elbo:e}:
% the full training ELBO produces an
% ``Occam hill''
% \cite{attias1999inferring,mackay1992bayesian}.
}
\label{fig:occam-hill}
\end{figure*}

For both the binary MNIST \cite{salakhutdinov2008quantitative}
and Omniglot \cite{lake2015omniglot} data sets,
we use a conventional convolutional architecture for the encoder and decoder, taken from \citet{pixelvae}.
The encoder is a convolutional neural network (CNN)
with 8 layers of kernel size 3, alternating strides of 1 and 2, and ReLU activations between the layers. The number of channels is 32 in the first 3 layers, then 64 in the last five layers.
Lastly, a linear layer transforms the CNN output to a vector of length $2 D I$, to produce the means $\bmu_{\bphi}(\x)$
and log variances $\log \bsigma^2_{\bphi}(\x)$ of
the posterior approximation of $\z$.
The decoder emits a Bernoulli likelihood
for every pixel independently.
Training details are provided in Appendix \ref{app:training-details}.

To demonstrate the benefits of using a structured prior, we ran experiments on MNIST and Omniglot using $I = 1$, with an increasing number of mixture components $K$.
(The data sets did not exhibit enough variability for additional factors $I > 1$ not to be pruned away \cite{mackay01local}).
Figure \ref{fig:occam-hill} shows the resulting test ELBO,
log-likelihood (negative reconstruction error) and KL-divergences from \eqref{eq:test-elbo}.

We decompose the ELBO in
\eqref{eq:test-elbo} into a log likelihood and two expected KL-divergences, $\KL[\z]$ and $\KL[\r]$.
In Figure \ref{fig:occam-hill} we see the ELBO increase,
and then seemingly converge,
as model complexity increases.
Following \eqref{eq:test-elbo},
the figure
shows that there is a trade-off in the penalty that the encoder pays, between encoding into $\z$, and encoding into $\r$.
As $K$ increases, $\KL[\r]$ increases and $\KL[\z]$ decreases, pushing information about the encoding ``higher up'' in the representation.
Beyond a certain $K$, proportionally less mixture components are used, reflected in a diminishing $\KL[\r]$.
We would expect Figure \ref{fig:occam-hill} to be smooth, and its variablility can be ascribed to
random seeds, sensible initialization of $q_{\blambda}(\bxi)$ and local minima in the $(\btheta, \bphi, \blambda)$ landscape of $\Lcal$.

An ELBO of -85.77 at $K=512$ on binarized MNIST
represents a typical state-of-the-art result ($\ge$ -87.4)
for unconditional generative models using simple convolutional neural networks \cite{pixelvae}.
On Omniglot, our best test ELBO of -90.89 at $K=256$
is comparable to some of the leading models of the day
(-89.8 for the variational lossy auto-encoder in \cite{chen2016variational};
$\ge$ -95.5 in \cite{rezende2016one};
$\ge$ -103.4 in \cite{burda2015importance}).

\begin{figure}[t]
\centering
\includegraphics[width=.055\textwidth]{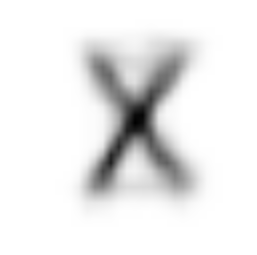}
\includegraphics[width=.055\textwidth]{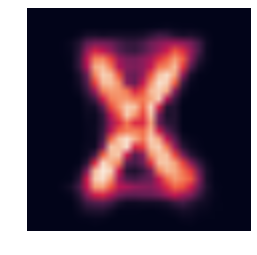}
\includegraphics[width=.055\textwidth]{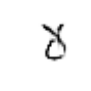}
\includegraphics[width=.055\textwidth]{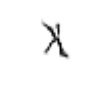}
\includegraphics[width=.055\textwidth]{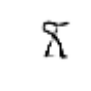}
\includegraphics[width=.055\textwidth]{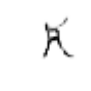}
\includegraphics[width=.055\textwidth]{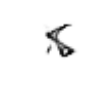}
\hspace{20pt}
\includegraphics[width=.055\textwidth]{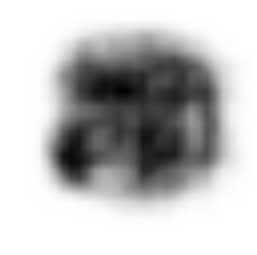}
\includegraphics[width=.055\textwidth]{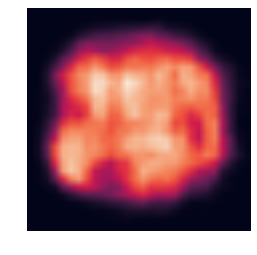}
\includegraphics[width=.055\textwidth]{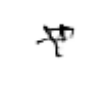}
\includegraphics[width=.055\textwidth]{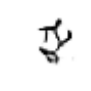}
\includegraphics[width=.055\textwidth]{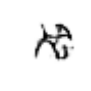}
\includegraphics[width=.055\textwidth]{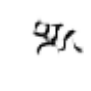}
\includegraphics[width=.055\textwidth]{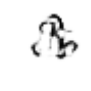} \\
\includegraphics[width=.055\textwidth]{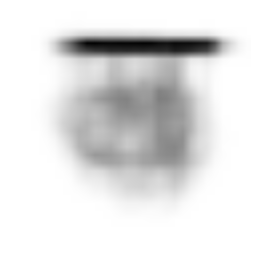}
\includegraphics[width=.055\textwidth]{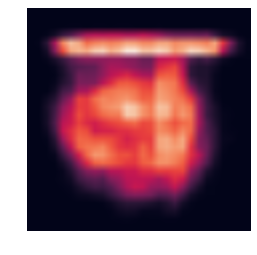}
\includegraphics[width=.055\textwidth]{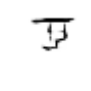}
\includegraphics[width=.055\textwidth]{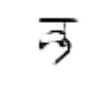}
\includegraphics[width=.055\textwidth]{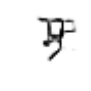}
\includegraphics[width=.055\textwidth]{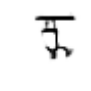}
\includegraphics[width=.055\textwidth]{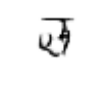}
\hspace{20pt}
\includegraphics[width=.055\textwidth]{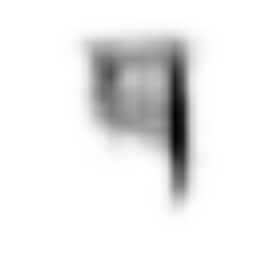}
\includegraphics[width=.055\textwidth]{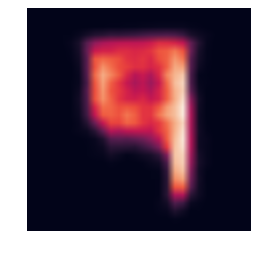}
\includegraphics[width=.055\textwidth]{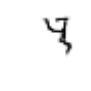}
\includegraphics[width=.055\textwidth]{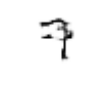}
\includegraphics[width=.055\textwidth]{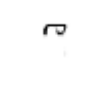}
\includegraphics[width=.055\textwidth]{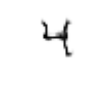}
\includegraphics[width=.055\textwidth]{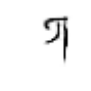} \\
\includegraphics[width=.055\textwidth]{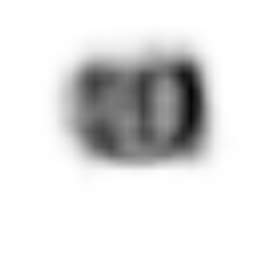}
\includegraphics[width=.055\textwidth]{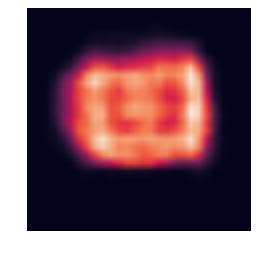}
\includegraphics[width=.055\textwidth]{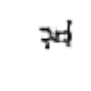}
\includegraphics[width=.055\textwidth]{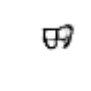}
\includegraphics[width=.055\textwidth]{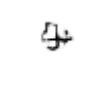}
\includegraphics[width=.055\textwidth]{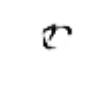}
\includegraphics[width=.055\textwidth]{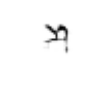}
\hspace{20pt}
\includegraphics[width=.055\textwidth]{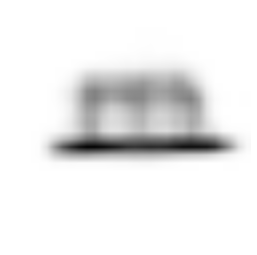}
\includegraphics[width=.055\textwidth]{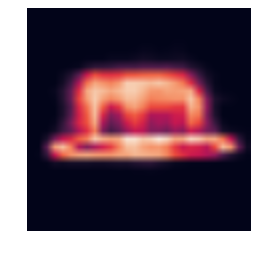}
\includegraphics[width=.055\textwidth]{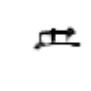}
\includegraphics[width=.055\textwidth]{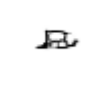}
\includegraphics[width=.055\textwidth]{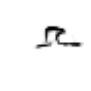}
\includegraphics[width=.055\textwidth]{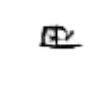}
\includegraphics[width=.055\textwidth]{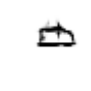} \\
\caption{Omniglot cluster means, variances, and samples (for 6 out of $K=32$ clusters).}
\label{fig:omniglot}
\vspace{-10pt}
\end{figure}

In Figure \ref{fig:omniglot}
we inspect how the latent space clustering of Omniglot is reflected in input space.
The figure shows the mean and variance
of samples of six mixture components, obtained by repeatedly drawing latent space samples, decoding those, and computing the mean and variance of the resulting images.
Additionally, five samples from each cluster are shown.

%!TeX root = main.tex

\subsection{CelebA}
\label{sec:celeba-results}

The images $\x^{\datapoint}$ in the CelebA dataset are 64-by-64 pixels with 3 channels (RGB),
with each element scaled to $[-1, 1]$.
For semi-supervised training,
we use the annotations of 40\% of images,
so that $|\Omega| = 0.4 N$.
Hence 40\% of data points
$\x^{\datapoint}$ have labels $\y^{\datapoint}$,
for which we use the subset of \texttt{gender},
\texttt{glasses} and \texttt{smiling} labels as a semi-supervised signal.
Our primary aim is to illustrate the factorial mixture prior capturing and encouraging compositional latent representations.

\begin{figure}[t]
\centering
\includegraphics[width=0.2\columnwidth]{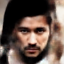} \phantom{0}
\includegraphics[width=0.2\columnwidth]{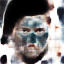} \phantom{0}
\includegraphics[width=0.2\columnwidth]{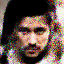}
\caption{The mean, standard deviation, and a $p_{\btheta}(\x | \z)$ sample.}
\label{fig:celeba-dist}
\vspace{-10pt}
\end{figure}

\begin{figure*}[t]
\centering
% \vskip 0.2in
\begin{tabular}[t]{c@{\hskip 4pt}c@{\hskip 4pt}c@{\hskip 4pt}c@{\hskip 4pt}c@{\hskip 4pt}c@{\hskip 4pt}c@{\hskip 4pt}c}
\includegraphics[width=0.1\textwidth]{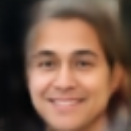} &
\includegraphics[width=0.1\textwidth]{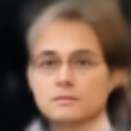} &
\includegraphics[width=0.1\textwidth]{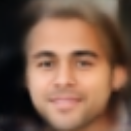} &
\includegraphics[width=0.1\textwidth]{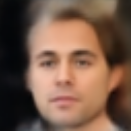} &
\includegraphics[width=0.1\textwidth]{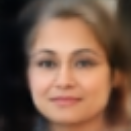} &
\includegraphics[width=0.1\textwidth]{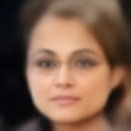} &
\includegraphics[width=0.1\textwidth]{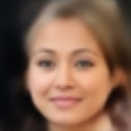} &
\includegraphics[width=0.1\textwidth]{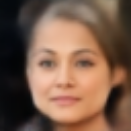} \\
\includegraphics[width=0.1\textwidth]{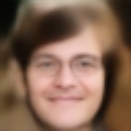} &
\includegraphics[width=0.1\textwidth]{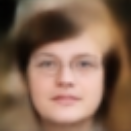} &
\includegraphics[width=0.1\textwidth]{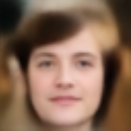} &
\includegraphics[width=0.1\textwidth]{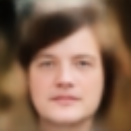} &
\includegraphics[width=0.1\textwidth]{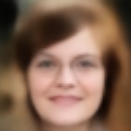} &
\includegraphics[width=0.1\textwidth]{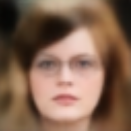} &
\includegraphics[width=0.1\textwidth]{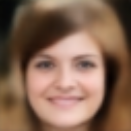} &
\includegraphics[width=0.1\textwidth]{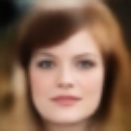} \\
\includegraphics[width=0.1\textwidth]{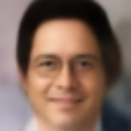} &
\includegraphics[width=0.1\textwidth]{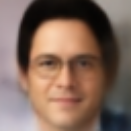} &
\includegraphics[width=0.1\textwidth]{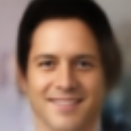} &
\includegraphics[width=0.1\textwidth]{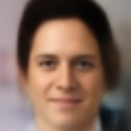} &
\includegraphics[width=0.1\textwidth]{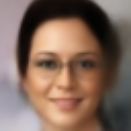} &
\includegraphics[width=0.1\textwidth]{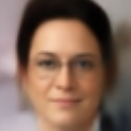} &
\includegraphics[width=0.1\textwidth]{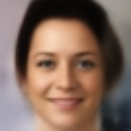} &
\includegraphics[width=0.1\textwidth]{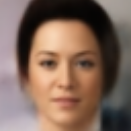} \\
$(\texttt{m}, \texttt{g}, \texttt{s})$ &
$(\texttt{m}, \texttt{g}, \texttt{ns})$ &
$(\texttt{m}, \texttt{ng}, \texttt{s})$ &
$(\texttt{m}, \texttt{ng}, \texttt{ns})$ &
$(\texttt{f}, \texttt{g}, \texttt{s})$ &
$(\texttt{f}, \texttt{g}, \texttt{ns})$ &
$(\texttt{f}, \texttt{ng}, \texttt{s})$ &
$(\texttt{f}, \texttt{ng}, \texttt{ns})$
\end{tabular}
\caption{
The means of $p_{\theta}(\x | \z)$, with $\z \sim p_{\blambda}(\z | k_1, k_2)$
sampled from \eqref{eq:z-t-dist}
using factorial mixture prior with
$I=5$ properties (samples from $p_{\theta}(\x | \z)$ add position-dependent jitter to the mean; see Figure \ref{fig:celeba-dist}). Semi-supervised labels
$k_1 \in \{\texttt{male}, \texttt{female} \}$,
$k_2 \in \{\texttt{glasses}, \texttt{no glasses} \}$, and
$k_3 \in \{\texttt{smiling}, \texttt{not smiling} \}$
are incorporated in learning, and the figure
shows randomly generated faces using the same set-up
as that of Figure \ref{fig:faces-and-glasses}.
}
\label{fig:faces-glasses-smiles}
\end{figure*}

For CelebA, we use a 4-layer convolutional neural network with (128, 256, 512, 1024) channels, kernel size 3, and stride 2.
The output of each layer is clipped with rectified linear units (ReLU), before being input to the next.
As with MNIST and Omniglot, the final layer output is
transformed with a linear layer to a vector of length $2 D I$, to produce the means $\bmu_{\bphi}(\x)$
and log variances $\log \bsigma^2_{\bphi}(\x)$ of $\z$.
The decoder is a deconvolutional neural network whose architecture is the transpose of that of the encoder. The decoder is different for this data set, as it needs to model 3 color channels (RGB).
The last layer of the deconvolutional network has 6 channels, which represent the mean and standard deviation
of $p_{\btheta}(\x | \z)$ for each color channel for each pixel; see Figure \ref{fig:celeba-dist}.
The standard deviations of $p_{\btheta}$ are parameterized via a scaled sigmoid function, to be in [0.001, 0.4].

Training started with a prior initialization phase for $\blambda$ of $2 \times 10^4$ iterations,
followed by a joint optimization phase of $3\times 10^5$ iterations.
The Adam optimizer \cite{kingma2014adam} with $10^{-4}$ learning rate was used for $\btheta$ and $\bphi$.
A forgetting rate $\kappa = -0.52$ and $\tau_0 = 2 \times 10^3$ was used for $\blambda$'s
gradient updates.
In training, we used $\Delta = 1000$ in \eqref{eq:semi-sup}, as a typical unscaled cross entropy loss in \eqref{eq:semi-sup} is a few orders of 
magnitude smaller than the reconstruction term in \eqref{eq:elbo:a}, which scales with
the number of pixels and color channels.
Empirically, we found that initially up-weighting the KL term in \eqref{eq:elbo:b} by a factor of 20 encouraged a crisper clustered representation to be learned \cite{higgins2017beta}.

Figure \ref{fig:faces-glasses-smiles} visually illustrates a model with
$I = 5$ factors with $D = 64$, of which
the first three are semi-supervised with two mixture components each,
while the remainder are unsupervised with $K_4 = K_5 = 64$.
For each row, we sample $(k_4, k_5)$ once from
$q_{\blambda}(\bxi)$ and sample
$[\z_4, \z_5]$ once form \eqref{eq:z-t-dist}.
By iterating over $(k_1, k_2, k_3)$ settings
and generating random faces from the model,
the change in properties is perceptible as the latent code changes.
Further results and examples are given in Appendix \ref{app:celeba}.

\begin{table}[t]
\vskip -5pt
\caption{Human judgements of properties of generated
$\x$'s, sampled according to
Section \ref{sec:sampling}
but with latent variables $k_1$ to $k_3$ clamped to intended, semi-supervised ground truth properties.
}
\label{tab:human-judgements}
\vskip -5pt
\begin{center}
\begin{tabular}{llc}
\toprule
& Latent property & Human judgement \\
\midrule
$k_1$ & \texttt{male} / \texttt{female}  &  94\% \\
$k_2$ & \texttt{glasses} / \texttt{no glasses}  & 85\% \\
$k_3$ & \texttt{smiling} / \texttt{not smiling}  & 81\% \\
\bottomrule
\end{tabular}
\end{center}
\vskip -10pt
\end{table}

To formally test the interpretability of samples like that in Figure \ref{fig:faces-glasses-smiles},
we uniformly sampled $(k_1, k_2, k_3)$ from $\{ 1, 2 \}^3$ as the model's ``intention'' to test all properties equally. It was appended with a $(k_4, k_5)$ sample from $q(\bpi)$, $\z | k_i$ was sampled according to \eqref{eq:z-t-dist} and the mean $p_{\btheta}(\x | \z)$ rendered.
For 54 such random images,
Table \ref{tab:human-judgements} evaluates shows the accuracy of human evaluations compared to their latent $k_i$-labels.
Ten people of varying race and gender evaluated the generated images.
The evaluation is subjective---the criteria for \texttt{smiling} need not be consistent with that of the creators of CelebA \cite{liu2015faceattributes}---but gives evidence
that the $k_i$'s capture interpretable attributes.

On the CelebA test set, the estimated cluster assignments $q_{\bphi \blambda}(\r_i | \x)$
were 97\% for \texttt{gender}, 99\% for the presence or absence of \texttt{glasses}, and 91\% for whether the subject is \texttt{smiling} or not.
Note that unlike the data evaluated for Table \ref{tab:human-judgements}, labeled properties are not uniformly distributed over the test set.

% \input{conclusion}

\section*{Acknowledgements}

We are grateful to three
anonymous Neural Information Processing Systems reviewers, who provided invaluable suggestions
for improving this paper.

\bibliography{mixture}

\begin{thebibliography}{35}
\providecommand{\natexlab}[1]{#1}
\providecommand{\url}[1]{\texttt{#1}}
\expandafter\ifx\csname urlstyle\endcsname\relax
  \providecommand{\doi}[1]{doi: #1}\else
  \providecommand{\doi}{doi: \begingroup \urlstyle{rm}\Url}\fi

\bibitem[Adel et~al.(2018)Adel, Ghahramani, and Weller]{adel18discovering}
Adel, T., Ghahramani, Z., and Weller, A.
\newblock Discovering interpretable representations for both deep generative
  and discriminative models.
\newblock In \emph{Proceedings of the 35th International Conference on Machine
  Learning}, pp.\  50--59, 2018.

\bibitem[Attias(1999)]{attias1999inferring}
Attias, H.
\newblock Inferring parameters and structure of latent variable models by
  variational {B}ayes.
\newblock In \emph{Proceedings of the Fifteenth Conference on Uncertainty in
  Artificial Intelligence}, pp.\  21--30, 1999.

\bibitem[Burda et~al.(2016)Burda, Grosse, and
  Salakhutdinov]{burda2015importance}
Burda, Y., Grosse, R., and Salakhutdinov, R.
\newblock Importance weighted autoencoders.
\newblock In \emph{International Conference on Learning Representations}, 2016.

\bibitem[Chen et~al.(2018{\natexlab{a}})Chen, Min, and Sun]{chen18kway}
Chen, T., Min, M.~R., and Sun, Y.
\newblock Learning k-way d-dimensional discrete codes for compact embedding
  representations.
\newblock In \emph{Proceedings of the 35th International Conference on Machine
  Learning}, pp.\  854--863, 2018{\natexlab{a}}.

\bibitem[Chen et~al.(2018{\natexlab{b}})Chen, Li, Grosse, and
  Duvenaud]{chen2018isolating}
Chen, T.~Q., Li, X., Grosse, R., and Duvenaud, D.
\newblock Isolating sources of disentanglement in variational autoencoders.
\newblock \emph{arXiv:1802.04942}, 2018{\natexlab{b}}.

\bibitem[Chen et~al.(2017)Chen, Kingma, Salimans, Duan, Dhariwal, Schulman,
  Sutskever, and Abbeel]{chen2016variational}
Chen, X., Kingma, D.~P., Salimans, T., Duan, Y., Dhariwal, P., Schulman, J.,
  Sutskever, I., and Abbeel, P.
\newblock Variational lossy autoencoder.
\newblock In \emph{International Conference on Learning Representations}, 2017.

\bibitem[Dilokthanakul et~al.(2016)Dilokthanakul, Mediano, Garnelo, Lee,
  Salimbeni, Arulkumaran, and Shanahan]{dilokthanakul16}
Dilokthanakul, N., Mediano, P.~A.~M., Garnelo, M., Lee, M.~C.~H., Salimbeni,
  H., Arulkumaran, K., and Shanahan, M.
\newblock Deep unsupervised clustering with {G}aussian mixture variational
  autoencoders.
\newblock \emph{arXiv:1611.02648}, 2016.

\bibitem[Fraccaro et~al.(2017)Fraccaro, Kamronn, Paquet, and
  Winther]{fraccaro17disentangled}
Fraccaro, M., Kamronn, S., Paquet, U., and Winther, O.
\newblock A disentangled recognition and nonlinear dynamics model for
  unsupervised learning.
\newblock In \emph{Advances in Neural Information Processing Systems 30}, pp.\
  3604--3613, 2017.

\bibitem[Ghahramani \& Beal(2000)Ghahramani and Beal]{beal2000variational}
Ghahramani, Z. and Beal, M.~J.
\newblock Variational inference for {B}ayesian mixtures of factor analysers.
\newblock In \emph{Advances in Neural Information Processing Systems 12}, pp.\
  449--455, 2000.

\bibitem[Graves et~al.(2018)Graves, Menick, and {van den
  Oord}]{graves2018associative}
Graves, A., Menick, J., and {van den Oord}, A.
\newblock Associative compression networks.
\newblock \emph{arXiv:1804.02476}, 2018.

\bibitem[Gulrajani et~al.(2017)Gulrajani, Kumar, Ahmed, Taiga, Visin, Vazquez,
  and Courville]{pixelvae}
Gulrajani, I., Kumar, K., Ahmed, F., Taiga, A.~A., Visin, F., Vazquez, D., and
  Courville, A.
\newblock Pixelvae: A latent variable model for natural images.
\newblock In \emph{International Conference on Learning Representations}, 2017.

\bibitem[Higgins et~al.(2017)Higgins, Matthey, Pal, Burgess, Glorot, Botvinick,
  Mohamed, and Lerchner]{higgins2017beta}
Higgins, I., Matthey, L., Pal, A., Burgess, C., Glorot, X., Botvinick, M.,
  Mohamed, S., and Lerchner, A.
\newblock Beta-{VAE}: Learning basic visual concepts with a constrained
  variational framework.
\newblock In \emph{International Conference on Learning Representations}, 2017.

\bibitem[Hoffman et~al.(2013)Hoffman, Blei, Wang, and
  Paisley]{hoffman2013stochastic}
Hoffman, M.~D., Blei, D.~M., Wang, C., and Paisley, J.
\newblock Stochastic variational inference.
\newblock \emph{Journal of Machine Learning Research}, 14\penalty0
  (1):\penalty0 1303--1347, 2013.

\bibitem[Jegou et~al.(2011)Jegou, Douze, and Schmid]{jegou2011product}
Jegou, H., Douze, M., and Schmid, C.
\newblock Product quantization for nearest neighbor search.
\newblock \emph{IEEE Transactions on Pattern Analysis and Machine
  Intelligence}, 33\penalty0 (1):\penalty0 117--128, 2011.

\bibitem[Jiang et~al.(2017)Jiang, Zheng, Tan, Tang, and
  Zhou]{jiang2016variational}
Jiang, Z., Zheng, Y., Tan, H., Tang, B., and Zhou, H.
\newblock Variational deep embedding: An unsupervised and generative approach
  to clustering.
\newblock In \emph{Proceedings of the 26th International Joint Conference on
  Artificial Intelligence}, pp.\  1965--1972, 2017.

\bibitem[Johnson et~al.(2016)Johnson, Duvenaud, Wiltschko, Adams, and
  Datta]{johnson2016svae}
Johnson, M., Duvenaud, D.~K., Wiltschko, A., Adams, R.~P., and Datta, S.~R.
\newblock Composing graphical models with neural networks for structured
  representations and fast inference.
\newblock In \emph{Advances in Neural Information Processing Systems 29}, pp.\
  2946--2954, 2016.

\bibitem[Kim \& Mnih(2018)Kim and Mnih]{kim2018disentangling}
Kim, H. and Mnih, A.
\newblock Disentangling by factorising.
\newblock \emph{arXiv:1802.05983}, 2018.

\bibitem[Kingma \& Ba(2015)Kingma and Ba]{kingma2014adam}
Kingma, D.~P. and Ba, J.
\newblock Adam: A method for stochastic optimization.
\newblock In \emph{International Conference on Learning Representations}, 2015.

\bibitem[Kingma \& Welling(2014)Kingma and Welling]{kingma2013auto}
Kingma, D.~P. and Welling, M.
\newblock Auto-encoding variational {B}ayes.
\newblock In \emph{International Conference on Learning Representations}, 2014.

\bibitem[Lake et~al.(2015)Lake, Salakhutdinov, and Tenenbaum]{lake2015omniglot}
Lake, B.~M., Salakhutdinov, R., and Tenenbaum, J.~B.
\newblock Human-level concept learning through probabilistic program induction.
\newblock \emph{Science}, 350\penalty0 (6266):\penalty0 1332--1338, 2015.

\bibitem[Lample et~al.(2017)Lample, Zeghidour, Usunier, Bordes, Denoyer, and
  Ranzato]{lample2017fader}
Lample, G., Zeghidour, N., Usunier, N., Bordes, A., Denoyer, L., and Ranzato,
  M.
\newblock Fader networks: Manipulating images by sliding attributes.
\newblock In \emph{Advances in Neural Information Processing Systems 30}, pp.\
  5967--5976. 2017.

\bibitem[Lin et~al.(2018)Lin, Hubacher, and Khan]{lin2018variational}
Lin, W., Hubacher, N., and Khan, M.~E.
\newblock Variational message passing with structured inference networks.
\newblock \emph{International Conference on Learning Representations}, 2018.

\bibitem[Liu et~al.(2015)Liu, Luo, Wang, and Tang]{liu2015faceattributes}
Liu, Z., Luo, P., Wang, X., and Tang, X.
\newblock Deep learning face attributes in the wild.
\newblock In \emph{Proceedings of International Conference on Computer Vision
  (ICCV)}, 2015.

\bibitem[MacKay(2001)]{mackay01local}
MacKay, D.~J.~C.
\newblock Local minima, symmetry-breaking, and model pruning in variational
  free energy minimization.
\newblock Technical report, University of Cambridge Cavendish Laboratory, 2001.

\bibitem[Nalisnick \& Smyth(2017)Nalisnick and Smyth]{nalisnick2017stick}
Nalisnick, E. and Smyth, P.
\newblock Stick-breaking variational autoencoders.
\newblock In \emph{International Conference on Learning Representations
  (ICLR)}, 2017.

\bibitem[Nalisnick et~al.(2016)Nalisnick, Hertel, and
  Smyth]{nalisnick2016approximate}
Nalisnick, E., Hertel, L., and Smyth, P.
\newblock Approximate inference for deep latent {G}aussian mixtures.
\newblock In \emph{NIPS Workshop on Bayesian Deep Learning}, volume~2, 2016.

\bibitem[Narayanaswamy et~al.(2017)Narayanaswamy, Paige, {van de Meent},
  Desmaison, Goodman, Kohli, Wood, and Torr]{narayanaswamy2017}
Narayanaswamy, S., Paige, T.~B., {van de Meent}, J.-W., Desmaison, A., Goodman,
  N., Kohli, P., Wood, F., and Torr, P.
\newblock Learning disentangled representations with semi-supervised deep
  generative models.
\newblock In \emph{Advances in Neural Information Processing Systems 30}, pp.\
  5925--5935. 2017.

\bibitem[Paquet(2015)]{paquet2015convergence}
Paquet, U.
\newblock On the convergence of stochastic variational inference in {B}ayesian
  networks.
\newblock \emph{arXiv:1507.04505}, 2015.

\bibitem[Paquet \& Koenigstein(2013)Paquet and Koenigstein]{paquet2013oneclass}
Paquet, U. and Koenigstein, N.
\newblock One-class collaborative filtering with random graphs.
\newblock In \emph{Proceedings of the 22nd International Conference on World
  Wide Web (WWW)}, pp.\  999--1008. 2013.

\bibitem[Rezende et~al.(2016)Rezende, Mohamed, Danihelka, Gregor, and
  Wierstra]{rezende2016one}
Rezende, D.~J., Mohamed, S., Danihelka, I., Gregor, K., and Wierstra, D.
\newblock One-shot generalization in deep generative models.
\newblock \emph{arXiv:1603.05106}, 2016.

\bibitem[Salakhutdinov \& Murray(2008)Salakhutdinov and
  Murray]{salakhutdinov2008quantitative}
Salakhutdinov, R. and Murray, I.
\newblock On the quantitative analysis of deep belief networks.
\newblock In \emph{Proceedings of the 25th International Conference on Machine
  Learning}, pp.\  872--879, 2008.

\bibitem[Tomczak \& Welling(2018)Tomczak and Welling]{tomczak2017vae}
Tomczak, J.~M. and Welling, M.
\newblock {VAE} with a {V}amp{P}rior.
\newblock In \emph{Proceedings of the 21st International Conference on
  Artificial Intelligence and Statistics (AISTATS)}, 2018.

\bibitem[{van den Oord} et~al.(2017){van den Oord}, Vinyals, and
  Kavukcuoglu]{vandenoord2017neural}
{van den Oord}, A., Vinyals, O., and Kavukcuoglu, K.
\newblock Neural discrete representation learning.
\newblock In \emph{Advances in Neural Information Processing Systems 30}, pp.\
  6306--6315. 2017.

\bibitem[Wainwright \& Jordan(2008)Wainwright and
  Jordan]{wainwright2008graphical}
Wainwright, M.~J. and Jordan, M.~I.
\newblock Graphical models, exponential families, and variational inference.
\newblock \emph{Foundations and Trends in Machine Learning}, 1\penalty0
  (1--2):\penalty0 1--305, 2008.

\bibitem[Waterhouse et~al.(1996)Waterhouse, MacKay, and
  Robinson]{waterhouse1996bayesian}
Waterhouse, S.~R., MacKay, D.~J.~C., and Robinson, A.~J.
\newblock Bayesian methods for mixtures of experts.
\newblock In \emph{Advances in Neural Information Processing Systems}, pp.\
  351--357, 1996.

\end{thebibliography}
\bibliographystyle{icml2018}

% \lipsum[5-10]
% \lipsum[5-10]

% \cleardoublepage
\clearpage
% \newpage

\appendix
%!TeX root = main.tex

%%%%%%%%%%%%%%%%%%%%%%%%%%%%%%%%%%%%%
\section{Evidence Lower Bound}
\label{app:elbo}
%%%%%%%%%%%%%%%%%%%%%%%%%%%%%%%%%%%%%

We bound the marginal likelihood of the entire data set---assuming $\x^{\datapoint}$'s are generated i.~i.~d.---with Jensen's inequality,
\begin{align*}
\log p(\X) & = \sum_{\R} \int p_{\btheta}(\X | \Z, \R) \, p(\R | \bxi) \, p(\bxi) \, \drm \Z \, \drm \bxi \\
& \ge \sum_{\R} \int q_{\bphi}(\Z | \X) \, q(\R) \, q(\bxi) \cdots \\
& \qquad\qquad \cdots \log
\Big[ p_{\btheta}(\X | \Z, \R) \, p(\R | \bxi) \, p(\bxi) \Big] \drm \Z \, \drm \bxi \\
& \quad - \sum_{\R} \int q_{\bphi}(\Z | \X) \, q(\R) \, q(\bxi) \cdots \\
& \qquad\qquad \cdots \log
\Big[ q_{\bphi}(\Z | \X) \, q(\R) \, q(\bxi) \Big] \drm \Z \, \drm \bxi \\
& \defined \Lcal(\btheta, \bphi, \blambda, \bgamma) \ .
\end{align*}
Using the joint distribution in \eqref{eq:joint},
and the approximation in \eqref{eq:approx}, we rewrite $\Lcal$ as
\begin{align*}
\Lcal & = \sum_n \Bigg\{ \Ebb_{q_{\bphi}(\z^{\datapoint} | \x^{\datapoint})} \left[ \log p_{\btheta}(\x^{\datapoint} | \z^{\datapoint}) \right] \\
& \qquad + \sum_i \Ebb_{q(\r_i^{\datapoint})} \Bigg[ \sum_k r_{ik}^{\datapoint} \Ebb_{q_{\bphi}(\z_i^{\datapoint} | \x^{\datapoint})} \cdots \\
& \qquad \cdots \Ebb_{q(\bmu_{ik}, \balpha_{ik}) \, q(\bpi_i)} \left[ 
\log p(\z_i^{\datapoint} | \bmu_{ik}, \balpha_{ik}) + \log \pi_{ik} \right] \Bigg] \\
& \qquad - \sum_i \Ebb_{q_{\bphi}(\z_i^{\datapoint} | \x^{\datapoint})} \left[ \log q_{\bphi}(\z_i^{\datapoint} | \x^{\datapoint}) \right] \\
& \qquad
 - \sum_i \Ebb_{q(\r_i^{\datapoint})} \left[ \log q(\r_i^{\datapoint}) \right] \\
& \qquad + \frac{1}{N} \sum_{i,k} \Ebb_{q(\bmu_{ik}, \balpha_{ik})} \left[ \log p(\bmu_{ik}, \balpha_{ik}) \right]
\\
& \qquad + \frac{1}{N} \sum_i \Ebb_{q(\bpi_i)} \left[ \log p(\bpi_i) \right] \\ 
& \qquad - \frac{1}{N} \sum_{i,k} \Ebb_{q(\bmu_{ik}, \balpha_{ik})}[ \log q(\bmu_{ik}, \balpha_{ik}) ] \\
& \qquad
- \frac{1}{N} \sum_i \Ebb_{q(\bpi_i)} \left[ \log q(\bpi_i) \right] \Bigg\} \ .
\end{align*}
Notice that we split the contribution of the priors equally over all data points through $\sum_n \frac{1}{N}$: this is simply so that we can run a stochastic gradient descent algorithm and appropriately weigh gradients with respect to 
$q(\bmu_{ik}, \balpha_{ik})$ in mini-batch steps.
Rearranging $\Lcal$
gives the ELBO in \eqref{eq:elbo:a} to \eqref{eq:elbo:d}.

\begin{figure}[t]
    \centering
    \includegraphics[width=\columnwidth]{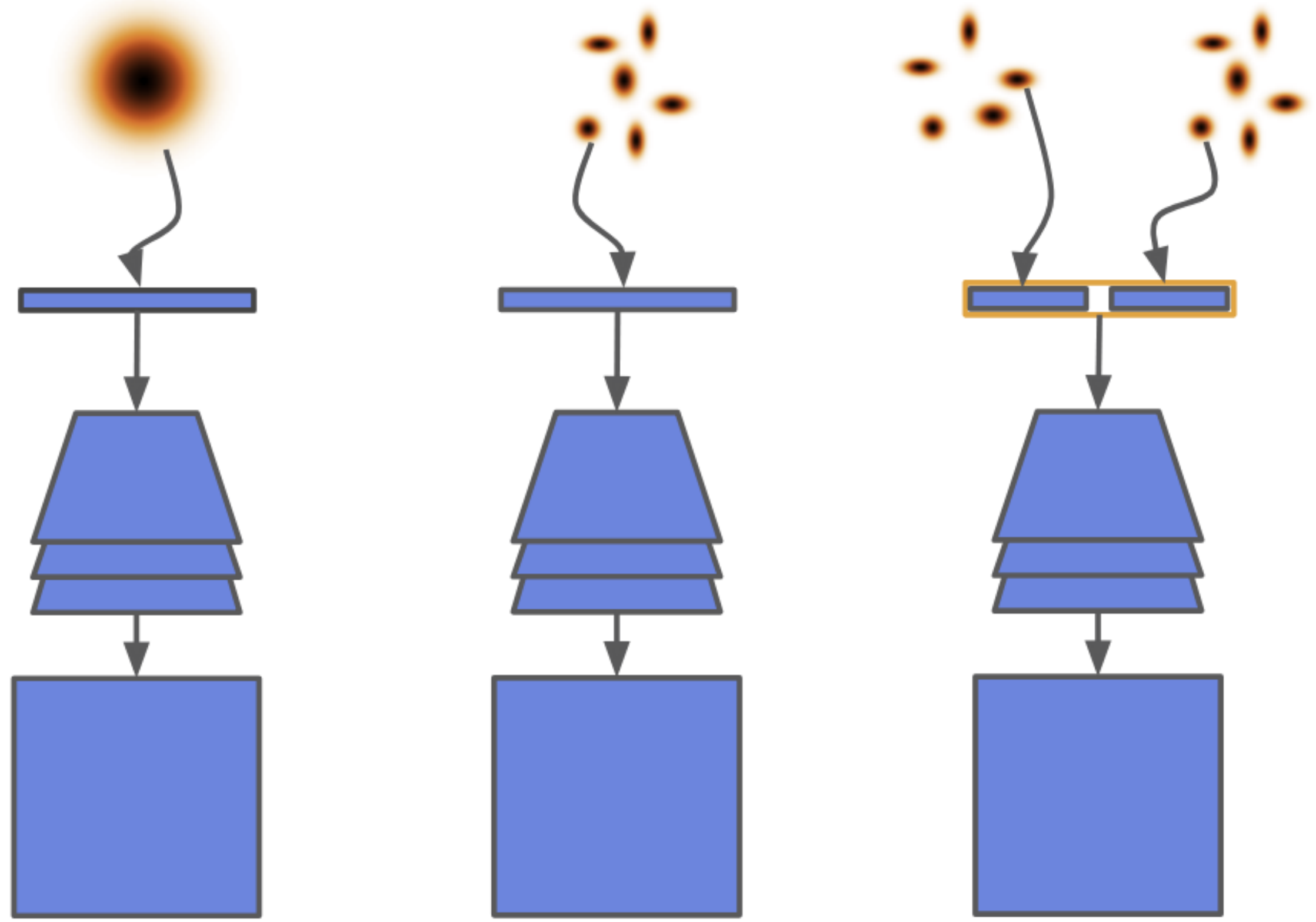}
    \caption{\emph{Left}, standard VAE generation. \emph{Center}, generation of $\x$ using a Gaussian mixture model prior. \emph{Right}, generation using a factorized structured prior.}
    \label{fig:model-diagram}
\end{figure}

The difference between a vanilla VAE and the model in this paper, is illustrated in Figure \ref{fig:model-diagram}.

%%%%%%%%%%%%%%%%%%%%%%%%%%%%%%%%%%%%%
\section{Mean and Natural Parameters}
\label{app:natural}
%%%%%%%%%%%%%%%%%%%%%%%%%%%%%%%%%%%%%

Approximation 
\eqref{eq:approx} contains factors
\begin{align*}
& q(\mu_{ikd}, \alpha_{ikd}) \\
& = \Ncal(\mu_{ikd} ; m_{ikd}, (s_{ikd} \alpha_{ikd})^{-1}) \, \Gcal(\alpha_{ikd} ; a_{ikd}, b_{ikd})
\end{align*}
which are stated in terms of their mean parameters
$\{  m_{ikd}, s_{ikd}, a_{ikd}, b_{ikd}\}$.
Instead of doing gradient descent in the mean parameterization, we will view 
$q_{\blambda}(\bxi)$ through the lens of its \emph{natural parameters} $\blambda$.
As an example of this lens, the Normal-Gamma distribution comprises of an inner product between natural parameter
vector $\blambda_{ikd}$ and sufficient statistics $\t(\mu_{ikd}, \alpha_{ikd})$,
\begin{align}
\log q(\mu, \alpha) & =
\big[ a - \tfrac{1}{2} ,  - (b + \tfrac{1}{2} s m^2), s m, - \tfrac{1}{2} s \big]^T \cdots \nonumber \\
& \qquad \cdots \big[\log \alpha, \alpha, \alpha \mu, \alpha \mu^2 \big] + \mathrm{const} \nonumber \\
& = \blambda_{ikd}^T \t(\mu_{ikd}, \alpha_{ikd}) + \mathrm{const} \label{eq:natural}
\end{align}
(dropping subscripts $ikd$ for brevity, and grouping constants).
The mapping from mean to natural parameters is therefore:
\begin{align*}
\lambda_{ikd, 1} & = a_{ikd} - \tfrac{1}{2} \\
\lambda_{ikd, 2} & = - (b_{ikd} + \tfrac{1}{2} s_{ikd} m_{ikd}^2) \\
\lambda_{ikd, 3} & = s_{ikd} m_{ikd} \\
\lambda_{ikd, 4} & = - \tfrac{1}{2} s_{ikd} \ .
\end{align*}

%%%%%%%%%%%%%%%%%%%%%%%%%%%%%%%%%%%%%%
\section{Kullback-Leibler Divergences}
\label{app:kl-divergences}
%%%%%%%%%%%%%%%%%%%%%%%%%%%%%%%%%%%%%%

%%%%%%%%%%%%%%%%%%%%%%%%%%%%%%%%%%%%%%%%%%%%
\subsection{KL Divergence Between Normal-Gamma Distributions}
\label{app:normal-gamma-kl}
%%%%%%%%%%%%%%%%%%%%%%%%%%%%%%%%%%%%%%%%%%%%

In \eqref{eq:elbo:d}, we are required to compute the KL divergence between two Normal-Gamma distributions.
We first state the prior Normal-Gamma distribution fully, for clarity:
\begin{align*}
& p(\bmu_{ik}, \balpha_{ik}) \\
& = \prod_d \Ncal(\mu_{ikd} ; m_0, (s_0 \alpha_{ikd})^{-1}) \, \Gcal(\alpha_{ikd} ; a_0, b_0) \\
& = \prod_d  \frac{b_0^{a_0}}{\Gamma(a_0)} \, \sqrt{\frac{s_0}{2 \pi}} \,
\exp \Bigg\{ \left( a_0 - \frac{1}{2} \right) \log \alpha_{ikd} \\
& \qquad\qquad - b_0 \alpha_{ikd} -\frac{1}{2} s_0 \alpha_{ikd} (\mu_{ikd} - m_{0})^2 \Bigg\} \\
& = \prod_d \frac{b_0^{a_0}}{\Gamma(a_0)} \, \sqrt{\frac{s_0}{2 \pi}} \,
\exp \Bigg\{
\left( a_0 - \frac{1}{2} \right) \underbrace{\log \alpha_{ikd}}_{*} \\
& \qquad\qquad
- \left(b_0 + \frac{1}{2} s_0 m_{0}^2 \right) \underbrace{\alpha_{ikd}}_{*} \\
& \qquad\qquad
+ s_0 m_{0} \underbrace{\alpha_{ikd} \mu_{ikd}}_{*}
- \frac{1}{2} s_0 \underbrace{\alpha_{ikd} \mu_{ikd}^2}_{*}
\Bigg\} \ .
\end{align*}
The sufficient statistics are indicated with a $*$,
and their moments are
\begin{align*}
\Ebb[\log \alpha_{ikd}] & = \psi(a_0) - \log b_0 \\
\Ebb[\alpha_{ikd}] & = \frac{a_0}{b_0} \\
\Ebb[\alpha_{ikd} \mu_{ikd}] & = m_0 \frac{a_0}{b_0} \\
\Ebb[\alpha_{ikd} \mu_{ikd}^2] & = \frac{1}{s_0} + m_0^2 \frac{a_0}{b_0} \ ,
\end{align*}
where $\psi(\cdot)$ indicates the digamma function.

The KL divergence, for one dimension $d$---where subscript $q$ denotes $q(\mu, \alpha)$'s parameters and subscript $0$ denotes the prior parameters---is
\begin{align*}
& \KL(q(\mu, \alpha) \| p(\mu, \alpha)) \\
& =
\frac{1}{2} \Bigg( s_0 \frac{a_q}{b_q} (m_q - m_0)^2
+ \frac{s_0}{s_q} - (\log(s_0) - \log(s_q)) - 1
\Bigg) \\
& \qquad + a_0  \log \left(\frac{b_q}{b_0} \right) 
- (\psi(a_q) - \psi(a_0)) \\
& \qquad
+ (a_q - a_0) \psi(a_q)
- (b_q - b_0) \frac{a_q}{b_q}
\ .
\end{align*}

%%%%%%%%%%%%%%%%%%%%%%%%%%%%%%%%%%%%%%%%%%%%%%
\subsection{Expected KL Divergence $\KL[\z_{i}]$}
%%%%%%%%%%%%%%%%%%%%%%%%%%%%%%%%%%%%%%%%%%%%%%

In \eqref{eq:elbo:b} and \eqref{eq:test-elbo}
we require
\begin{align*}
\KL[\z_{i}] 
& =
\Ebb_{q_{\blambda}(\bmu, \balpha)} \Bigg[
\sum_{k = 1}^{K_i}
\Ebb_{q} [r_{ik}] \, \cdots \\
& \qquad \cdots \KL \Big(
q_{\bphi}(\z_i | \x) \, \Big\| \,
p(\z_i | \bmu_{ik}, \balpha_{ik}) \Big) \Bigg]
\\
& = 
\Ebb_{q_{\blambda}(\bmu, \balpha)} \Bigg[
- \frac{1}{2} \sum_{k = 1}^{K_i} \gamma_{ik}
\sum_{d = 1}^D \Bigg( \cdots
\\
& \qquad \log \bsigma_{\bphi}^2(\x)_{d'} + \log \alpha_{ikd'} + 1
\\
& \qquad - \alpha_{ikd'} \Big(\bsigma_{\bphi}^2(\x)_{d'} + 
(\bmu_{\bphi}(\x)_{d'} - \mu_{ikd'})^2 \Big) 
\Bigg)\Bigg]
\\
& = 
- \frac{1}{2} \sum_k \gamma_{ik}
\sum_{d} \Bigg( \cdots
\\
& \qquad \log \bsigma_{\bphi}^2(\x)_{d'} + \psi(a_{ikd'}) - \log b_{ikd'}
\\
& \qquad - \frac{a_{ikd'}}{b_{ikd'}} \Big(\bsigma_{\bphi}^2(\x)_{d'} + 
(\bmu_{\bphi}(\x)_{d'} - m_{ikd'})^2 \Big) \\
& \qquad - \frac{1}{s_{ikd'}} + 1
\Bigg) \ . 
\end{align*}
Indices $d' = d + (i - 1)D$ index the correct elements of $\bmu_{\bphi}(\x)$
and $\bsigma_{\bphi}^2(\x)$.

%%%%%%%%%%%%%%%%%%%%%%%%%%%%%%%%%%%%%%%%%%%%
\subsection{Expected KL Divergence $\KL[\r_i]$}
%%%%%%%%%%%%%%%%%%%%%%%%%%%%%%%%%%%%%%%%%%%%

In \eqref{eq:elbo:c} and \eqref{eq:test-elbo}
we require
\begin{align*}
\KL[\r_i] & =
\Ebb_{q_{\blambda}(\bpi_i)} \left[ \KL \Big( q_{\bgamma}(\r_i) \, \Big\| \, p(\r_i | \bpi_i) \Big) \right]
\\
& = 
\Ebb_{q_{\blambda}(\bpi_i)} \left[ \sum_k \gamma_{ik} (\log \gamma_{ik}
- \log \pi_{ik}) \right] \\
& = 
\sum_k \gamma_{ik} \left( \log \gamma_{ik} + \psi\left(\sum_{k'} c_{ik'} \right) - \psi(c_{ik}) \right) \ .
\end{align*}

%%%%%%%%%%%%%%%%%%%%%%%%%%%
\section{Natural Gradients}
\label{app:natural-gradients}
%%%%%%%%%%%%%%%%%%%%%%%%%%

We wish to update the mixture component parameters $\m_{ik}$, $\s_{ik}$, $\a_{ik}$ and $\b_{ik}$, and other \emph{mean} parameters.
We will write the gradient step in terms of
\emph{natural} parameters.
In particular, the stochastic ELBO for minibatch $\Bcal$
with respect to $\blambda_{ik}$ is
\begin{align*}
& \Lcal_{\Bcal}(\blambda_{ik}) = \\
& \Ebb_{q(\bmu_{ik}, \balpha_{ik})} \Bigg[ \frac{N}{|\Bcal|} \sum_{n \in \Bcal} \gamma_{ik}^{\datapoint}
\Ebb_{q_{\bphi_i}(\z_i^{\datapoint} | \x^{\datapoint})}
\big[ \log p(\z_i^{\datapoint} | \bmu_{ik}, \balpha_{ik}) \big] \\
&  \qquad \; +  \log p(\bmu_{ik}, \balpha_{ik}) - \log q(\bmu_{ik}, \balpha_{ik}) \Bigg] \ .
\end{align*}
Expanding the inner terms yields
\begin{align*}
& \Lcal_{\Bcal}(\blambda_{ik}) = \\
& \sum_{d} \Ebb_{q(\mu_{ikd}, \alpha_{ikd})}
\Bigg[ \frac{N}{|\Bcal|} \sum_{n \in \Bcal} \gamma_{ik}^{\datapoint}
\Bigg( \\
& \qquad\qquad \frac{1}{2} \underbrace{\log \alpha_{ikd}}_{*} - \frac{1}{2} \log (2 \pi) \\
& \qquad\qquad
- \frac{1}{2} \Big(\bmu_{\bphi}(\x^{\datapoint})_{d'}^2 + \bsigma_{\bphi}^2(\x^{\datapoint})_{d'} \Big) \underbrace{\alpha_{ikd}}_{*} \\
& \qquad\qquad
+ \bmu_{\bphi}(\x^{\datapoint})_{d'} \underbrace{\alpha_{ikd} \mu_{ikd}}_{*}
- \frac{1}{2} \underbrace{\alpha_{ikd} \mu_{ikd}^2}_{*}
\Bigg) \\
& \qquad
+ \log \left( \frac{b_0^{a_0}}{\Gamma(a_0)} \, \sqrt{\frac{s_0}{2 \pi}} \right)
+ \left( a_0 - \frac{1}{2} \right) \underbrace{\log \alpha_{ikd}}_{*}
\\
& \qquad - \left(b_0 + \frac{1}{2} s_0 m_{0}^2 \right) \underbrace{\alpha_{ikd}}_{*} \\
& \qquad
+ s_0 m_{0} \underbrace{\alpha_{ikd} \mu_{ikd}}_{*} - \frac{1}{2} s_0 \underbrace{\alpha_{ikd} \mu_{ikd}^2}_{*} \Bigg]
\\
&
- \sum_{d} \Ebb_{q(\mu_{ikd}, \alpha_{ikd})} \Big[ \log q(\mu_{ikd}, \alpha_{ikd}) \Big] \ ,
\end{align*}
where we have arranged the terms matching the natural parameters and sufficient statistics, which we indicate with a $*$, in
the same order as in \eqref{eq:natural}.
Indices $d' = d + (i - 1)D$ index the correct elements of $\bmu_{\bphi}(\x^{\datapoint})$
and $\bsigma_{\bphi}^2(\x^{\datapoint})$,
which we conveniently wrote over terms matching the
natural parameters and sufficient
statistics in \eqref{eq:natural}.
We regroup terms:
\begin{align*}
& \Lcal_{\Bcal}(\blambda_{ik}) = \\
& \sum_{d} \Ebb_{q(\mu_{ikd}, \alpha_{ikd})} \Bigg[
\left( \left( a_0 + \frac{1}{2} \frac{N}{|\Bcal|} \sum_{n \in \Bcal}
\gamma_{ik}^{\datapoint} \right) - \frac{1}{2} \right) \underbrace{\log \alpha_{ikd}}_{*} \\
& \qquad
- \Bigg(b_0 + \frac{1}{2} s_0 m_{0}^2 + \frac{1}{2} \frac{N}{|\Bcal|} \sum_{n \in \Bcal} \, \gamma_{ik}^{\datapoint}
\cdots \\
& \qquad\qquad\qquad \cdots
\Big(\bmu_{\bphi}(\x^{\datapoint})_{d'}^2 + \bsigma_{\bphi}^2(\x^{\datapoint})_{d'} \Big) \Bigg) \underbrace{\alpha_{ikd}}_{*} \\
& \qquad
+ \left(s_0 m_{0} + \frac{N}{|\Bcal|} \sum_{n \in \Bcal} \gamma_{ik}^{\datapoint}
\bmu_{\bphi}(\x^{\datapoint})_{d'} \right) \underbrace{\alpha_{ikd} \mu_{ikd}}_{*} \\
& \qquad
- \frac{1}{2} \left( s_0 + \frac{N}{|\Bcal|} \sum_{n \in \Bcal} \gamma_{ik}^{\datapoint} \right) \underbrace{\alpha_{ikd} \mu_{ikd}^2}_{*} \Bigg] \\
&
- \sum_{d} \Ebb_{q(\mu_{ikd}, \alpha_{ikd})} \Big[ \log q(\mu_{ikd}, \alpha_{ikd}) \Big]
+ \mathrm{const} \ .
\end{align*}
For mini-batch $\Bcal$ of data points,
$\Lcal_{\Bcal}(\blambda_{ik})$ is minimized with respect at
$\blambda_{ikd}$, as natural parameters in the form of \eqref{eq:natural}:
\begin{align}
\lambda_{ikd, 1}^*
& = a_0 + \frac{1}{2} \frac{N}{|\Bcal|} \sum_{n \in \Bcal} \gamma_{ik}^{\datapoint} - \frac{1}{2} \nonumber \\
\lambda_{ikd, 2}^*
& =  - \Bigg( b_0 + \frac{1}{2} s_0 m_{0}^2 \nonumber \\
& \qquad + \frac{1}{2} \frac{N}{|\Bcal|} \sum_{n \in \Bcal} \gamma_{ik}^{\datapoint}
\Big(\bmu_{\bphi}(\x^{\datapoint})_{d'}^2 + \bsigma_{\bphi}^2(\x^{\datapoint})_{d'} \Big) \Bigg) \nonumber \\
\lambda_{ikd, 3}^*
& = 
\left(s_0 m_{0} + \frac{N}{|\Bcal|} \sum_{n \in \Bcal} \gamma_{ik}^{\datapoint} \bmu_{\bphi}(\x^{\datapoint})_{d'} \right)
\nonumber \\
\lambda_{ikd, 4}^*
& = - \frac{1}{2} \left( s_0 + \frac{N}{|\Bcal|} \sum_{n \in \Bcal} \gamma_{ik}^{\datapoint} \right) \ .
\label{eq:natural-gradients}
\end{align}
The values of $\blambda_{ikd}^*$ in \eqref{eq:natural-gradients}
constitute the natural gradients.

At iteration $\tau$, we determine $\blambda^{(\tau)}$ from iteration $\tau$'s mean parameters of $q_{\blambda}(\bxi)$,
using the form in \eqref{eq:natural}. Using $\blambda^*$ in \eqref{eq:natural-gradients}, we update $\blambda$
using the convex combination
\begin{equation} \label{eq:lambda-update}
\blambda^{(\tau + 1)} \leftarrow (1 - \rho_{\tau}) \blambda^{(\tau)} + \rho_{\tau} \blambda^* \ .
\end{equation}
This parameter update is a rewritten from of updating 
$\blambda^{(\tau)}$ with the natural gradient $\widehat{\nabla}_{\blambda^{(\tau)}} \Lcal(\blambda)$
that is rescaled by $\rho_{\tau}$ \cite{hoffman2013stochastic,paquet2015convergence}.
The step sizes should satisfy $\sum_{\tau = 1}^{\infty} \rho_{\tau} = \infty$ and $\sum_{\tau = 1}^{\infty} \rho_{\tau}^2 < \infty$.
In our results $\rho_{\tau} = (\tau_0 + \tau)^{-\kappa}$ was used,
with forgetting rate $\kappa \in (\frac{1}{2}, 1]$
and delay $\tau_0 \ge 0$.
The updated \emph{mean parameters} of $q(\bxi)$ are obtained from the updated natural parameters $\blambda^{(\tau + 1)}$.
The Dirichlet pseudo-counts of $q(\bpi_i)$ are similarly updated with
$c_{ik}^{(\tau + 1)} \leftarrow (1 - \rho_{\tau}) c_{ik}^{(\tau)} + \rho_{\tau} (c_0 + \frac{N}{|\Bcal|} \sum_{n \in \Bcal} \gamma_{ik}^{\datapoint})$.

%%%%%%%%%%%%%%%%%%%%%%%%%%
\section{Training details}
\label{app:training-details}
%%%%%%%%%%%%%%%%%%%%%%%%%%

All models were trained on a P100 GPU with 16 GB of memory.
Only the CelebA models required that amount of memory; the MNIST and Omniglot models needed significantly less on account of their lower numbers of parameters and smaller activation vectors.

Pre-training (see Section \ref{sec:inference}),
when used, lasted for $10^5$ iterations in all cases, followed by $2\times10^4$ prior initialization iterations.
The batch size was constant at 64 across runs, phases and models.

\subsection{MNIST and Omniglot}

The MNIST and Omniglot models all reached convergence after $2\times10^5$ iterations of joint optimization; the complete runs took approximately 4 hours.

The MNIST and Omniglot models had 567,775 parameters in $\btheta$ and $\bphi$ taken together, and for the prior another (latent dimension $\times$ 3 + 1) $\times$ number of components.
For a model with 64 latent dimensions and 64 prior components, that adds to 12,352 extra parameters, leading to a total model size of 580,127.

\subsection{CelebA}

The CelebA runs took approximately $3\times10^5$ iterations to converge, for a total training time of about 14 hours. 
he CelebA iterations took longer because they required more computation, and more data fetching:
data points are approximately $16$ times larger than for MNIST and Omniglot.

The CelebA model had 27,857,670 parameters between them
$\btheta$ and $\bphi$,
and the prior approximation for the best run an additional 49,280 parameters in $\blambda$, for a total of 27,906,050.

\subsection{Optimization}

All models were trained using the Adam optimizer \cite{kingma2014adam}, with a learning rate of $10^{-4}$ for the embedding and generation parameters $\btheta$ and $\bphi$, while a forgetting rate $\kappa = -0.7$ and $\tau_0 = 2 \times 10^3$ was used for $\blambda$'s
gradient updates. The initial Robbins-Monro fraction was $0.01$ in all cases.

%%%%%%%%%%%%%%%%%
\section{CelebA}
\label{app:celeba}
%%%%%%%%%%%%%%%%%

\begin{figure}[t]
\centering
% \ifdefined\bigfig
\includegraphics[width=\columnwidth]{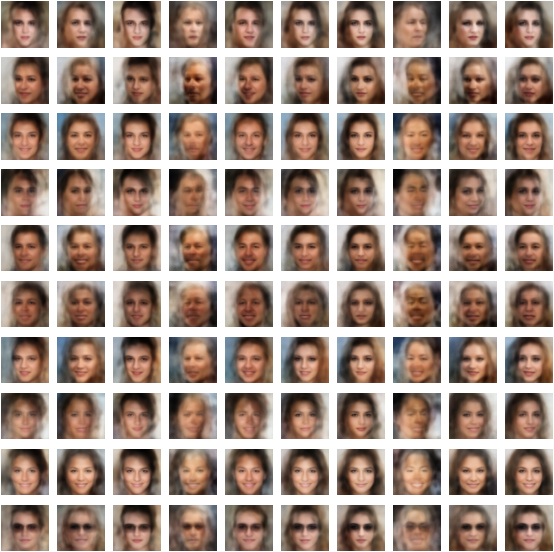}
% \else
% figure here
% \fi
\caption{
The means of samples for $\x$, generated from $p_{\btheta}(\x | \z) \, p_{\blambda}(\z | k_1, k_2)$ as $k_1$ and $k_2$ varies along the rows and columns.
The figure shows a
$10 \times 10$ subset of a $64 \times 64$ grid.
The model was trained wholly unsupervised, with no KL weights (like what would appear in models
based on the $\beta$-VAE set-up), and the clustering structure is a consequence of the Bayesian hierarchical prior.
We may discern image properties that some settings of $k_1$ and $k_2$ are responsible for modeling;
the bottom row's $k_1$ captures a ``glasses'' cluster, even though no supervised labels were ever presented.
Some of the mixture component approximations $q_{\blambda}(\bpi_i)$
had pseudo-counts very close to that of the prior $c_0$ for some index values of $k_i$ and were effectively pruned; this explains the very ``washed'' column (third left).
}
\label{fig:unsupervised}
\end{figure}

In Figure \ref{fig:unsupervised} we show 
the average of 100 conditional samples for $\x$ from
$p_{\btheta}(\x | \z) p_{\blambda}(\z | k_1, k_2)$
where $p_{\blambda}(\z | k_1, k_2)$ is sampled according to \eqref{eq:z-t-dist}.
The model has $I=2$, $D=128$, $K_1 = K_2 = 64$, and is trained in a completely unsupervised way.
There were no KL weights (like what would appear in models
based on the $\beta$-VAE set-up), and the clustering structure is a consequence of the Bayesian hierarchical prior.
We iterate over all $(k_1, k_2)$ pairs to create the grid of images, and the figure shows a
$10 \times 10$ subset of the $64 \times 64$ grid.
Specific visual properties of the clusters can be detected in the rows and columns. 

\begin{figure*}[h]
\centering
\begin{subfigure}[b]{0.485\textwidth}
\centering
%\ifdefined\bigfig
\includegraphics[width=\textwidth]{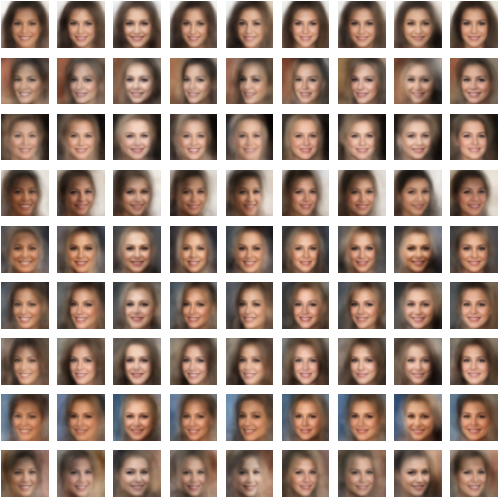}
%\else figure here \fi
\caption{$(\texttt{female}, \texttt{no glasses}, \texttt{smiling})$}
\end{subfigure}%
$\phantom{00}$
\begin{subfigure}[b]{0.485\textwidth}
\centering
 %\ifdefined\bigfig
\includegraphics[width=\textwidth]{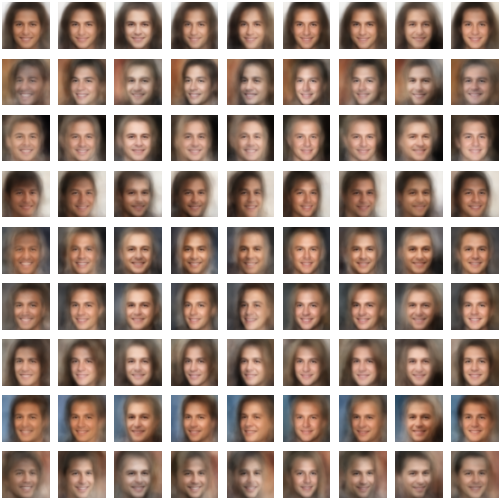}
% \else figure here \fi
\caption{$(\texttt{male}, \texttt{no glasses}, \texttt{smiling})$}
\end{subfigure}
\\
\vspace{8pt}
\begin{subfigure}[b]{0.485\textwidth}
\centering
%\ifdefined\bigfig
\includegraphics[width=\textwidth]{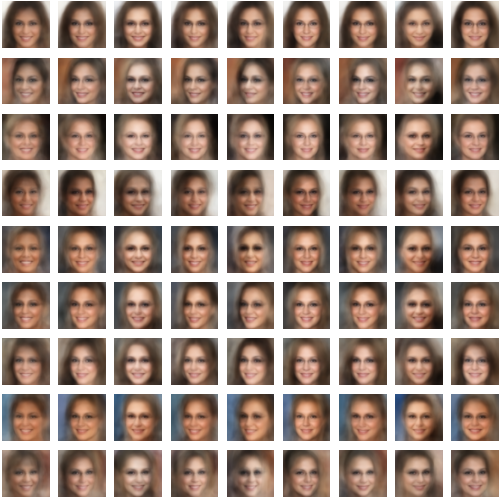}
%\else figure here \fi
\caption{$(\texttt{female}, \texttt{glasses}, \texttt{smiling})$}
\end{subfigure}%
$\phantom{00}$
\begin{subfigure}[b]{0.485\textwidth}
\centering
% \ifdefined\bigfig
\includegraphics[width=\textwidth]{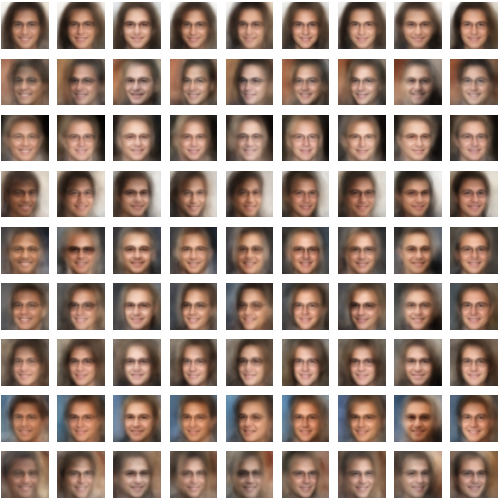}
% \else figure here \fi
\caption{$(\texttt{male}, \texttt{glasses}, \texttt{smiling})$}
\end{subfigure}%
\caption{
The means of samples for $\x$, generated from $p_{\btheta}(\x | \z) \, p_{\blambda}(\z | k_1, k_2, k_3, k_4, k_5)$ as $k_4$ and $k_5$ varies along the rows and columns.
The figure shows a
$9 \times 9$ subset of a $64 \times 64$ grid.
Mixture component indices $(k_1, k_2)$ are set to their four settings, to which we associated
semi-supervised properties. Mixture component index $k_3 = 1$
is set to its associated property ``\texttt{smiling}''.
Sub-figure titles show their semi-supervised $(k_1, k_2, k_3)$ settings.
}
\label{fig:celeba-smiling}
\end{figure*}

\begin{figure*}[h]
\centering
\begin{subfigure}[b]{0.485\textwidth}
\centering
% \ifdefined\bigfig
\includegraphics[width=\textwidth]{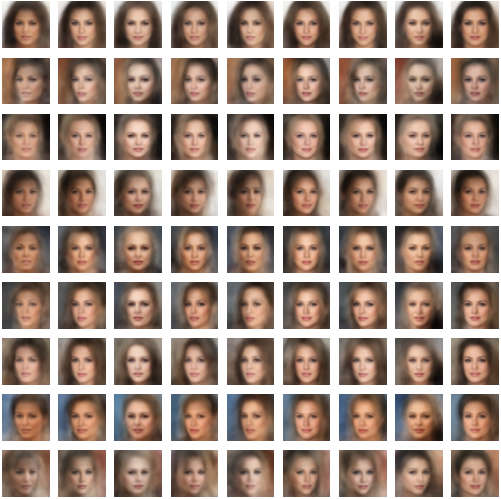}
% \else figure here \fi
\caption{$(\texttt{female}, \texttt{no glasses}, \texttt{not smiling})$}
\end{subfigure}%
$\phantom{00}$
\begin{subfigure}[b]{0.485\textwidth}
\centering
% \ifdefined\bigfig
\includegraphics[width=\textwidth]{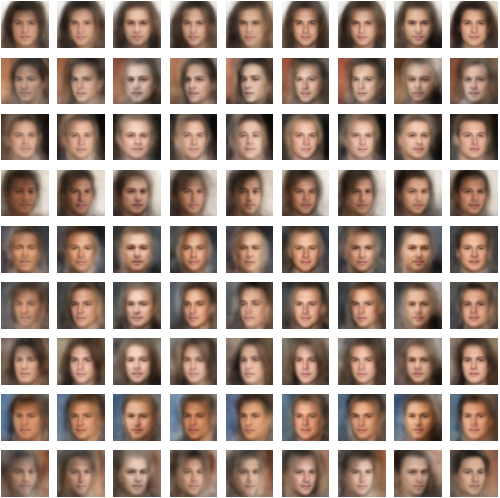}
% \else figure here \fi
\caption{$(\texttt{male}, \texttt{no glasses}, \texttt{not smiling})$}
\end{subfigure}
\\
\vspace{8pt}
\begin{subfigure}[b]{0.485\textwidth}
\centering
% \ifdefined\bigfig
\includegraphics[width=\textwidth]{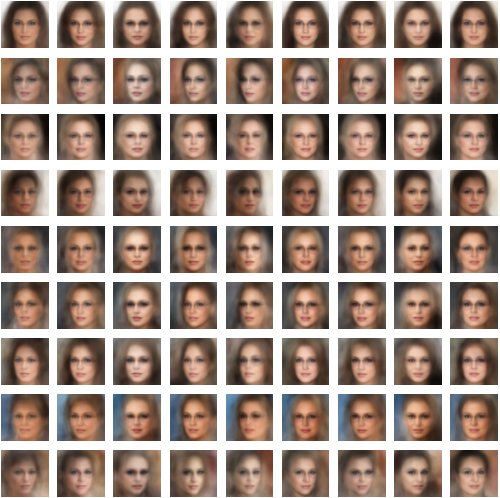}
% \else figure here \fi
\caption{$(\texttt{female}, \texttt{glasses}, \texttt{not smiling})$}
\end{subfigure}%
$\phantom{00}$
\begin{subfigure}[b]{0.485\textwidth}
\centering
% \ifdefined\bigfig
\includegraphics[width=\textwidth]{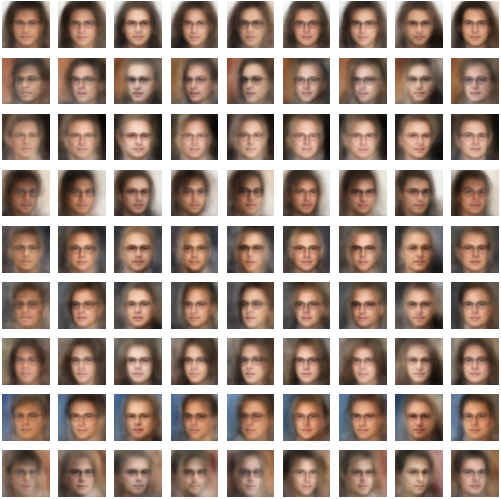}
% \else figure here \fi
\caption{$(\texttt{male}, \texttt{glasses}, \texttt{not smiling})$}
\end{subfigure}%
\caption{
The means of samples for $\x$, generated from $p_{\btheta}(\x | \z) \, p_{\blambda}(\z | k_1, k_2, k_3, k_4, k_5)$ as $k_4$ and $k_5$ varies along the rows and columns.
The figure shows a
$9 \times 9$ subset of a $64 \times 64$ grid.
Mixture component indices $(k_1, k_2)$ are set to their four settings, to which we associated
semi-supervised properties. Mixture component index $k_3 = 2$
is set to its associated property ``\texttt{not smiling}''.
Sub-figure titles show their semi-supervised $(k_1, k_2, k_3)$ settings.
Compare Figure \ref{fig:celeba-smiling}.
}
\label{fig:celeba-not-smiling}
\end{figure*}

In the eight sub-figures in
Figures \ref{fig:celeba-smiling} and \ref{fig:celeba-not-smiling} we illustrate the eight settings
of $(k_1, k_2, k_3)$ for the model in Section
\ref{sec:celeba-results}.
Each sub-figure shows 
a $9 \times 9$ sub-grids of the $64 \times 64$ grid,
iterating settings $k_4 = 1, .., 64$ and $k_2 = 1, .., 64$ across the rows and columns.
Each face is an average of 100 conditional samples for $\x$ from
$p_{\btheta}(\x | \z) \, p_{\blambda}(\z | k_1, k_2, k_3, k_4, k_5)$
for the respective $k_i$-settings.

\clearpage

\clearpage

\end{document}